\pdfoutput=1
\documentclass[11pt]{article}

\usepackage[preprint]{acl}

\usepackage{times}
\usepackage{latexsym}
\usepackage{subcaption}
\usepackage{multirow}
\usepackage{tabularx}
\usepackage{tcolorbox}
\usepackage{makecell}

\usepackage{tikz}
\usetikzlibrary{positioning, fit, backgrounds}
\tcbuselibrary{breakable}
\usetikzlibrary{arrows.meta}

\newtcolorbox{promptbox}[1][]{
    breakable,
    colback=gray!10,
    colframe=gray!50,
    fonttitle=\bfseries,
    title=#1
}

\usepackage[T1]{fontenc}

\usepackage[utf8]{inputenc}

\usepackage{microtype}

\usepackage{inconsolata}

\usepackage{graphicx}

\usepackage{lipsum}
\usepackage{subcaption, mdframed}
\usepackage{amsmath}
\usepackage{amssymb}
\usepackage[normalem]{ulem}

\newcommand{\conf}[1]{%
  \ifcase#1
  \or
    ($\uparrow$)%
  \or
    ($\uparrow\uparrow$)%
  \or
    ($\uparrow\uparrow\uparrow$)%
  \else
    (#1$\uparrow$)%
  \fi
}

\usepackage{booktabs}

\title{Legal Experts Disagree With Rationale Extraction Techniques for Explaining ECtHR Case Outcome Classification}

\author{
 \textbf{Mahammad Namazov\textsuperscript{1}},
 \textbf{Tomáš Koref\textsuperscript{2}},
 \textbf{Ivan Habernal\textsuperscript{1}}
\\
\\
 \textsuperscript{1}Trustworthy Human Language Technologies, Research Center Trustworthy Data Science \\and Security of the University Alliance Ruhr, Ruhr University Bochum,\\
 \textsuperscript{2}Center for Critical Computational Studies, Goethe University
Frankfurt
\\
 \small{
   \textbf{Correspondence:}\{\href{mailto:mahammad.namazov@ruhr-uni-bochum.de}{mahammad.namazov}, \href{ivan.habernal@ruhr-uni-bochum.de}{ivan.habenal}\}@ruhr-uni-bochum.de},\\
 }

\begin{document}
\maketitle
\begin{abstract}

Interpretability is critical for applications of large language models (LLMs) in the legal domain, where trust and transparency are essential. A central NLP task in this setting is legal outcome prediction, where models forecast whether a court will find a violation of a given right. We study this task on decisions from the European Court of Human Rights (ECtHR), introducing a new ECtHR dataset with carefully curated positive (violation) and negative (non‑violation) cases. Existing works propose both task-specific approaches and model-agnostic techniques to explain downstream performance, but it remains unclear which techniques best explain legal outcome prediction. To address this, we propose a comparative analysis framework for model-agnostic interpretability methods. We focus on two rationale extraction techniques that justify model outputs with concise, human-interpretable text fragments from the input. We evaluate faithfulness via normalized sufficiency and comprehensiveness metrics, and plausibility via legal expert judgments of the extracted rationales. We also assess the feasibility of using LLM-as-a-Judge, using these expert evaluations as reference. Our experiments on the new ECtHR dataset show that models’ “reasons” for predicting violations differ substantially from those of legal experts, despite strong faithfulness scores. The source code of our experiments is publicly available at \url{https://github.com/trusthlt/IntEval.}
\end{abstract}

\section{Introduction} \label{sec: introduction}
To what extent can we trust a classification prediction of a state-of-the-art model in the legal domain? While recent advancements in many LLMs applications focus on improving the performance, they come at the cost of interpretability \cite{lipton_mythos_2018, rudin_stop_2019, esposito_transparency_2022}.\footnote{We use the term interpretability, see the discussion regarding explainability and interpretability in Section \ref{sec: litrev}.} To interpret the decision-making process of complex models, various techniques have been proposed in the legal domain, such as task-specific techniques (i.e., techniques that mimic legal decision making processes) \cite{zhong_legal_2018, zhong_iteratively_2020, luo_prototype-based_2023} and model-agnostic techniques (i.e., methods which apply to any architecture) \cite{staliunaite_comparative_2024, chalkidis_paragraph-level_2021}. 
Although task-specific methods might provide valuable solutions for explaining complex legal reasoning, by design, they defy comparative analysis across tasks. Therefore, model-agnostic techniques are more suitable for a comparative evaluation. 

Among these, rationale extraction techniques, which explain the model's output using text fragments (i.e., rationales) from the input text, have been adopted in various setups \cite{chalkidis_paragraph-level_2021, staliunaite_comparative_2024}. However, whether or not they are actually useful has rarely been scrutinized by legal experts.
Moreover, an expert analysis is expensive and does not scale. To bypass this limitation, some studies investigate the qualitative analysis capabilities of LLMs \cite{matter_close_2024, mirzakhmedova_are_2024}, but LLM-as-a-Judge has not been studied to evaluate the interpretability techniques in the legal domain. In this paper, we ask the following research questions:
\begin{table*}[]
\centering
\begin{tabular}{l|rrr|rrr|rrr}
\toprule
\multicolumn{1}{l}{} &\multicolumn{3}{c}{\thead{\textbf{Reported non-violations} \\ \textbf{(including wrong labels)}}} & 
\multicolumn{3}{c}{\thead{\textbf{Actual} \\ \textbf{non-violations}}} & 
\multicolumn{3}{c}{\thead{\textbf{Directly/Partially} \\ \textbf{inadmissible cases}}} \\
\textbf{Dataset} & Train & Test & Dev. & Train & Test & Dev. & Train & Test & Dev. \\
\midrule
\citet{chalkidis_neural_2019} & 3549& 1024& 690& 652& 382& 109& 2612/24& 631/5& 527/3\\
\citet{chalkidis_paragraph-level_2021} & 762& 135& 148& 726& 132& 147& 2/7& 0/1& 0/0\\
\bottomrule
\end{tabular}
\caption{Gold-standard errors in previously published ECtHR datasets.}
\label{tab:stats_previous}
\end{table*}
\begin{itemize}
    \item Do legal experts find model-agnostic rationale extraction techniques capable of explaining violation predictions on ECtHR decisions?
    \item To what extent can we substitute costly human expert evaluation of extracted rationales with automated methods using LLMs?
\end{itemize}

To address this challenge, we develop a standardized comparative analysis framework to assess the interpretability of rationale-extraction techniques. Unlike existing works, we propose a schema for post-hoc evaluation of extracted rationales by legal experts. In addition, we evaluate rationale extraction techniques quantitatively (i.e., via a faithfulness evaluation).

Furthermore, we introduce a new ECtHR dataset to address limitations in existing binary violation existence classification benchmarks. We found previous datasets from \citet{chalkidis_neural_2019} and \citet{chalkidis_paragraph-level_2021} contain factual legal errors in their gold-labeled data for violation existence classification due to two critical issues. Both \emph{inadmissible cases} (those that do not even pass the requirements to be decided on the merits by the ECtHR) and cases with \textit{"implicit violations"} (i.e., the conclusion of the case does not match hand-crafted keywords, such as "violation")
were wrongly gold-labeled as non-violation. When we exclude such cases, the number of \textit{true non-violated} cases shrinks drastically (see Table~\ref{tab:stats_previous}).


We adopt two model-agnostic interpretability techniques, each representing a major family of rationale extraction methods: an optimization-based approach \cite{brinner_model_2023} and an ensemble-based approach \cite{chrysostomou_flexible_2021}. We evaluate both using normalized sufficiency, normalized comprehensiveness, and changes in F1 under input masking (see Section~\ref{sec: compareval}), and we compare their performance against the individual feature-attribution methods that make up the ISR ensemble \cite{chrysostomou_flexible_2021}.

To complement these quantitative faithfulness metrics, legal experts analyze the extracted rationales using global criteria that consider the full case context, in contrast to prior work that often studies rationales in isolation. Our expert analysis focuses on cases involving a single article violation.
Finally, we use the expert judgments as a reference point to investigate the capabilities of LLM-as-a-Judge for evaluating rationale quality.

\section{Related work} \label{sec: litrev}

While some researchers differentiate between interpretability and explainability terms \cite{rudin_stop_2019, bhambhoria_interpretable_2022}, others use them interchangeably \cite{jain_learning_2020, jacovi_towards_2020}, as we do here.

\citet{lipton_mythos_2018} and \citet{esposito_transparency_2022} discuss the distinction between transparency and explanations based on the focus of interpretability techniques. Studies that aim to build white-box models are either inherently interpretable \cite{rudin_stop_2019} or utilize domain-specific decision-making processes \cite{zhong_legal_2018, luo_prototype-based_2023, hu_ella_2024}. Furthermore, \citet{esposito_transparency_2022} argues that decisions, rather than the internal processes of machines, must be explained. Researchers explain the decisions of black-box models using post-hoc methods \cite{chrysostomou_flexible_2021, brinner_model_2023} to extract meaningful information from the inputs. For instance, \citet{sundararajan_axiomatic_2017} use gradients of the prediction with respect to the input, while \citet{jain_learning_2020} use attention weights to extract rationales. \citet{chrysostomou_flexible_2021} argue that the techniques for scoring token-level significance for rationale extraction must be chosen flexibly. However, their method extracts only a single rationale per text, which is unsuitable for longer legal documents that may contain multiple rationales.

\citet{chalkidis_paragraph-level_2021} propose an optimization technique to extract paragraph-level rationales from ECtHR decisions. \citet{brinner_model_2023} propose a similar approach that optimizes an input mask to extract rationales satisfying sufficiency, comprehensiveness, and compactness constraints \cite{yu_rethinking_2019}. Another challenge in interpretability applications is evaluation of such techniques. \citet{mohseni_multidisciplinary_2021} provide a ``catalogue'' and a guideline for evaluating XAI techniques. The work demonstrates that there is no single, solid evaluation method applicable to all techniques, which creates ambiguity, as argued by \citet{jacovi_towards_2020} in the context of faithfulness evaluation. According to \citet{jacovi_towards_2020}, defining interpretation in an ad hoc manner fails to distinguish between the evaluation of faithfulness, plausibility, and readability. 

\citet{deyoung_eraser_2020} propose the ERASER benchmark for the quantitative assessment of rationale extraction techniques. \citet{nguyen_comparing_2018} compares human and automatic evaluations of local explanations using a forward prediction task \cite{doshi-velez_towards_2017}, where human annotators must predict the outcome from the rationales alone. In the legal setting, \citet{chalkidis_paragraph-level_2021} use 50 expert-annotated `gold' rationales, together with silver rationales, to assess the quality of the extracted rationales. Although \citet{staliunaite_comparative_2024} provide a qualitative analysis, they do not involve legal experts, which weakens the credibility of their evaluation. 


Scholars increasingly use LLMs as evaluators (“LLM-as-a-Judge”) for qualitative analyses \cite{mirzakhmedova_are_2024, parfenova_text_2025}. \citet{bojic_comparing_2025} show that LLMs can match human inter-rater reliability in sentiment and political leaning tasks. Similarly, \citet{matter_close_2024} find that agreement between LLMs and humans is comparable to agreement among human annotators, and \citet{mirzakhmedova_are_2024} report high consistency between LLM- and human-generated annotations for most argument quality dimensions. However, to the best of our knowledge, LLM-as-a-Judge has not yet been used to evaluate interpretability techniques in the legal domain.

\section {Interpretability methods} \label{sec: methods_new}
We evaluate the capabilities of two interpretability methods, namely \textbf{Ma}sked \textbf{R}ationale \textbf{C}reation\footnote{Notice that, this is a rationale extraction technique even though \citet{brinner_model_2023} used the term of \emph{Creation} in the name.} \cite{brinner_model_2023} and Flexible \textbf{I}nstance \textbf{S}pecific \textbf{R}ationale Extraction \cite{chrysostomou_flexible_2021}, in the violation existence classification task on ECtHR decisions. 

\subsection{Masked Rationale Creation (MaRC)}\citet{brinner_model_2023} introduce MaRC, an interpretability method that optimizes an input mask in a supervised manner to extract meaningful, class-indicative fragments—i.e., evidence that supports the ground-truth label—from the input text. The mask is learned to replace non‑indicative parts of the input with uninformative tokens (e.g., \texttt{<PAD>}), while preserving the model’s class prediction based on the remaining tokens. In contrast to \citet{brinner_model_2023} whose method relied on the classification gold-label (e.g., violation/non-violation) as an `oracle' label for rationale extraction, we take a realistic scenario and use the predicted classification label instead. We argue that a deployed model can only rely on predictions during inference, and has never access to gold standard labels. MaRC formulates rationale extraction as an optimization problem with a regularizer that enforces three constraints proposed by \citet{yu_rethinking_2019}: 
\begin{itemize} 
    \item Rationales must be \textbf{sufficient} to replace the input text, such that classification performance remains similar to that obtained with the original input. 
    \item Rationales must be \textbf{comprehensive}, such that removing them substantially degrades downstream task performance. 
    \item Rationales must be \textbf{compact}, such that they are as short as possible while still serving as effective replacements for the input text. 
\end{itemize} By regularizing mask optimization with these constraints, the authors study the extraction of rationales that are simultaneously sufficient, comprehensive, and compact. For the mathematical details of the optimization technique and our modifications, we refer the reader to Appendix~\ref{sec: app_config}.

\subsection{Flexible Instance-Specific Rationale Extraction (ISR)} \label{subsec: isr}
ISR is a flexible rationale extraction method that constructs candidate masks by combining various feature-attribution methods—including random scoring, attention weights ($\alpha$) \cite{jain_learning_2020}, scaled attention ($\alpha \nabla \alpha$) \cite{serrano_is_2019}, integrated gradients ($\mathrm{IG}$) \cite{sundararajan_axiomatic_2017}, InputXGrad ($\mathbf{x}\nabla\mathbf{x}$) \cite{kindermans_investigating_2016, atanasova_diagnostic_2020}, DeepLift \cite{shrikumar_learning_2017}, and LIME \cite{ribeiro_why_2016}—with two rationale types (\emph{top-K} and \emph{contiguous-K}) and different rationale lengths (see Appendix \ref{sec: app_config} for details on scoring methods). ISR employs Jensen–Shannon divergence (JSD)\footnote{\citet{chrysostomou_flexible_2021} also consider Kullback-Leibler divergence \cite{kullback1951information}, perplexity \cite{jelinek1977perplexity} and class difference metrics, but we use JSD, as it yields the best performance according to their experiments.} \cite{lin1991divergence} to compute the divergence between the model's predictions on the masked input and on the original input. As a result, for each input sample ISR evaluates every combination of feature-attribution method, rationale type and rationale length, and selects as the rationale the combination whose masked input yields the highest divergence from the original input.

\citet{chrysostomou_flexible_2021} argue that a one-size-fits-all approach is not a robust solution, since rationale length can vary across input samples. ISR addresses this issue by collecting all possible rationales of length $l$, where $l \in [1, \dots, N]$, where $N$ is the maximum length of a single rationale which is predefined to 2.5\% of the entire sample length. 

In our legal-domain setting, we set the minimum rationale length to five tokens, as it is very unlikely that a single token constitutes a meaningful rationale. While the original ISR formulation allows single-token rationales \cite{chrysostomou_flexible_2021}, we prevent rationales from being single-token by enforcing this minimum length. Furthermore, we configure ISR to extract multiple rationales from the input text, unlike the original setup, since a legal decision can rarely be made on the basis of a single rationale. Finally, we restrict the rationale type to be contiguous rather than selecting top-K tokens. We provide further details on these modifications and on the ensemble structure of ISR in Section~\ref{sec: app_config}.

\section{Downstream legal task} \label{sec: downstream}We use legal decisions from the ECtHR to define a binary classification task based on the existence of a violation. For our experiments, we construct a new ECtHR dataset by scraping the HUDOC\footnote{\url{https://hudoc.echr.coe.int/}} database\footnote{The new ECtHR dataset is publicly available at \url{huggingface.co/datasets/TrustHLT/ECtHR_dataset}}. We collaborate with an additional legal expert, distinct from the four‑expert team that developed the evaluation guidelines (see Section \ref{subsec: qualysis}), to specify the filters used to create the dataset (see Appendix \ref{sec: app_data}).

\subsection{ECtHR dataset} 
\begin{table}
    \centering
    \begin{tabular}{l|rr}
        \toprule
        \textbf{} & \textbf{ECtHR} & \textbf{ECtHR-subs.} \\
        \midrule
        Positive cases &  15,221 & 1,573 \\
        Negative cases & 1,573 & 1,573 \\
        Tokens & 43.3M & 9.4M \\
        Tokens / doc &2.8k & 4.3k \\
        \bottomrule
    \end{tabular}
    \caption{Statistical details for the datasets; ECtHR-subs. denotes the balanced subset used for experiments in this paper.}
    \label{tab: dataset_statistics}
\end{table}

As we already highlighted in Section~\ref{sec: introduction}, existing datasets by \citet{chalkidis_neural_2019, chalkidis_paragraph-level_2021} contain a large amount of wrongly assigned gold labels for the category non-violation. To fix this issue, we scrape the HUDOC database up to December 2025 and obtain approximately 390k cases. After discarding cases with corrupted metadata and non-English documents, we retain 89k documents. We apply filtering keywords obtained by the legal expert to distinguish negative and positive instances for dataset construction. 

Using this filtering process we collect 1,573 cases that do not contain a violation of human rights. We retain all cases with at least one violated article, including those with implicit violations, as positive samples. Furthermore, we remove cases in which the same article is found both violated and non-violated, since specific provisions can simultaneously satisfy and infringe article requirements (e.g., Rusishvili vs. Georgia\footnote{\url{https://hudoc.echr.coe.int/eng?i=001-218078}}). In total, we obtain 15,221 cases in which ECtHR finds a violation of fundamental rights. We refer the reader to Appendix \ref{sec: app_data} for further details on the dataset creation process.

To balance the dataset for this study, we subsample violation cases from the new ECtHR dataset to match the number of non-violation cases (see Table~\ref{tab: dataset_statistics}). In this subsampling process, we preserve both the article distribution, especially Articles~6 (right to fair trial) and~8 (right to respect for private and family life) of European Convention of Human Rights\footnote{ Convention for the Protection of Human Rights and Fundamental Freedoms, see  \url{https://www.echr.coe.int/documents/d/echr/convention_ENG}} (ECHR), which are the most frequently violated provisions in the corpus, and the temporal distribution for negative and positive samples. 

\subsection{Downstream task: Violation detection as binary document classification}
We employ LEGAL-BERT \cite{chalkidis-etal-2020-legal}\footnote{We also experimented with Legal-Longformer \cite{mamakas-etal-2022-processing} and Modern-BERT \cite{modernbert}, but LEGAL-BERT achieved better downstream classification performance.} in a hierarchical configuration for the classification task. Following \citet{brinner_model_2023}, we split each sample into chunks to fit the 512-token limit of the classifier. Unlike MaRC's random per-instance chunk sampling to represent a document, we use cross-attention across chunks—following the HIER-BERT \citep{chalkidis_neural_2019} architecture—to preserve inter-chunk information flow for document-level classification. 

\section{Comparative evaluation of rationale extraction methods} \label{sec: compareval}
We evaluate the rationale extraction techniques using both quantitative metrics and expert analysis. We then use the expert evaluation results to assess the abilities of three LLMs to judge rationale quality in an LLM-as-a-Judge setting.
\subsection{Quantitative analysis}
We assess the interpretability techniques using normalized sufficiency (NormSuff) and normalized comprehensiveness (NormComp) metrics that were also employed by \citet{chrysostomou_flexible_2021}. In addition, we evaluate each technique by measuring the change in F1-score under sufficiency and comprehensiveness settings.

\paragraph{Normalized sufficiency}
We measure sufficiency as introduced by \citet{carton-etal-2020-evaluating} to quantify how well rationales support the model's original prediction. Specifically, sufficiency compares the classifier's confidence in its predicted label $\hat{y}$ for the full input $\mathbf{x}$ with its confidence when only the extracted rationales $\mathcal{R}$ are provided. For sufficiently informative rationales, this difference is expected to be small.
\begin{equation}
\text{Suff}(\mathbf{x}, \hat{y}, \mathcal{R}) = 1 - \max(0, p(\hat{y}|\mathbf{x}) - p(\hat{y}|\mathcal{R}))
\label{eq: suff}
\end{equation}
However, following \citet{carton-etal-2020-evaluating} we bound this difference between 0 and 1 by reversing the computation of sufficiency and clipping negative differences at 0. Hence, higher values indicate more sufficient rationales. To obtain the normalized sufficiency, we use an uninformative input—where all text content is masked out—as a baseline, i.e., $\text{Suff}(\mathbf{x}, \hat{y}, 0)$.
\begin{equation}
\text{NormSuff}(\mathbf{x}, \hat{y}, \mathcal{R}) = \frac{\text{Suff}(\mathbf{x}, \hat{y}, \mathcal{R}) - \text{Suff}(\mathbf{x}, \hat{y}, 0)}{1 - \text{Suff}(\mathbf{x}, \hat{y}, 0)}
\label{eq: norm_suff}
\end{equation}
\paragraph{Normalized comprehensiveness} We assess comprehensiveness using the NormComp metric \cite{carton-etal-2020-evaluating}. The metric computes comprehensiveness by measuring the change in prediction of the classifier when the rationales are removed from the input, i.e., when it is evaluated on $\mathbf{x}_{\textbackslash\mathcal{R}}$. Larger values indicate that the rationales contain important information for the model's prediction.
\begin{equation}
    \text{Comp}(\mathbf{x}, \hat{y}, \mathcal{R}) = \text{max}(0, p(\hat{y}|\mathbf{x}) - p(\hat{y}|\mathbf{x}_{\textbackslash\mathcal{R}}))
    \label{eq: comp}
\end{equation}
As it was done for sufficiency analysis, we use the uninformative input as a baseline ($\text{Comp}(\mathbf{x}, \hat{y}, 0)$).
\begin{equation}
    \text{NormComp}(\mathbf{x}, \hat{y}, \mathcal{R}) = \frac{\text{Comp}(\mathbf{x}, \hat{y}, \mathcal{R})}{1 - \text{Suff}(\mathbf{x}, \hat{y}, 0)}
    \label{eq: norm_comp}
\end{equation}

\paragraph{F1 score}
To assess the impact of rationale extraction on classifier performance, we measure macro averaged F1 scores under two settings: 
\begin{itemize}
\item Suff-F1: We compute the F1 score using only the extracted rationales as input. Higher value indicates that rationales alone suffice for accurate performance.
\item Comp-F1: We measure F1 score with the input where all rationales are masked out \cite{arras2016explaining}. In contrast to the Suff-F1, lower values indicate better rationale extraction.
\end{itemize}

\subsection{Qualitative analysis} \label{subsec: qualysis}
\subsubsection{Expert analysis} \label{subsec: expert_eval}

\begin{figure}\centering
    \includegraphics[width=0.8\linewidth]{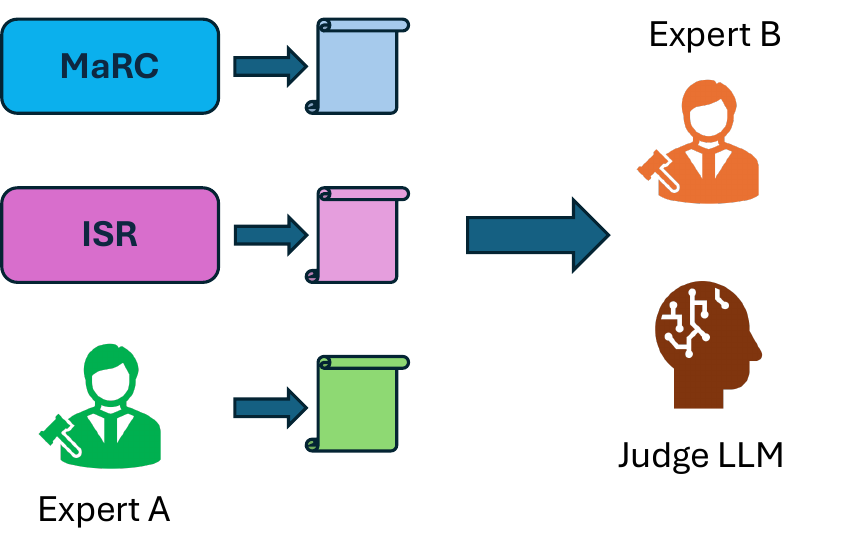}
    \caption{The scheme of the qualitative analysis, Expert A and the techniques extract rationales, which are then evaluated by Expert B and the LLMs.}
    \label{fig: eval}
\end{figure}

\paragraph{Task formulation} Our expert evaluation consists of two steps (Figure~\ref{fig: eval}). First, Expert~A extracts rationales from ten ECtHR cases that contain only the facts of the cases, and involve either Article~6 or Article~8 (i.e., five documents per article). Second, Expert~B assesses whether the rationales support the outcome and are sufficient to decide on the outcome. Expert~B also evaluates the rationales extracted by the MaRC and ISR techniques.
\paragraph{Annotation guidelines} Expert~A reviews only the facts of the case, without access to the Court's written reasoning or conclusion, but knowing both the article at issue and that the Court found a violation of that article \cite{chalkidis_paragraph-level_2021}. Expert~A then extracts the information they believe support the outcome.

The guidelines for Expert~B's evaluation were developed as part of this work by a team of four legal experts. Legal experts first labeled three pilot decisions independently, met  to resolve disagreements, and revised the guidelines. This iterative process repeated for four, two, and three more cases, so twelve pilot decisions in total shaped the final evaluation guidelines. Because the existing literature offers little guidance on how to evaluate legal rationales, disagreements were examined in depth, and the team agreed on categories that reflect human rights reasoning and existing scholarship (e.g., defining more precisely when a rationale “supports” a conclusion). For further details on annotation guidelines and the training phase, we refer the reader to Appendix~\ref{sec: app_ratex}. 

Expert B evaluates rationales from MaRC, ISR, and Expert A, all on the same ten ECtHR cases, resulting in 30 documents in total.
\paragraph{Global rationale support}
We provide Expert~B with the set of highlighted rationales for each case. The evaluation task is to decide whether provided these rationales, taken together, \emph{support} the existence of violation. 
\paragraph{Global rationale sufficiency}
For each legal decision, Expert~B uses the corresponding set of rationales to assess whether they are \emph{sufficient} to deduce the existence of a violation, or whether more information is required.

For both tasks, Expert~B sees only the rationales, without any additional context. We select ECtHR cases that Experts~A and~B were unfamiliar with.

\subsubsection{LLM-as-a-Judge} \label{subsec: llmjudge}
While expert evaluation provides the gold standard, it does not scale to large datasets due to time and financial costs. Motivated by recent work showing that LLMs can be used to evaluate model performance \cite{parfenova_text_2025, mirzakhmedova_are_2024}, we test their capabilities in the legal domain. Specifically we compare LLMs against Expert~B's assessments of rationales with respect to support and sufficiency criteria.

We employ Llama-3.1 (70B) \cite{grattafiori2024llama}, Mixtral-8$\times$7B \cite{jiang2024mixtral}, and Gemma-2-27B-it \cite{team2024gemma} to evaluate rationales from ten unique legal cases across three sources (MaRC, ISR and expert A).\footnote{We also employed SaulLM \cite{colombo2024saullm} in LLM-as-a-Judge setting. However, technical limitations of the model hinder us to get the adequate performance (see Appendix \ref{sec: app_prompts}).} We set the temperature to 0.0, as the task requires determinism rather than creative generation \cite{renze2024effect}.

We run our experiments in both zero-shot and few-shot configurations. In the few-shot setup, we use a leave-one-out approach within each article (Article~6 or~8): expert assessments for four documents are used as examples to judge the fifth, ensuring no cross-article contamination. We report Cohen's Kappa scores \cite{cohen1960coefficient} to quantify inter-annotator agreement between each LLM's rationale assessments and the expert evaluations. In addition we compute bootstrapped confidence intervals (CI) to account for the limited number of documents.
\begin{table*}[]
    \centering
    \begin{tabular}{l|rrrrrrr|rr}
        \toprule
        \textbf{Technique}&\textbf{Random}&\textbf{LIME}&\textbf{DeepLift}&\textbf{IG}&$\mathbf{x}\nabla\mathbf{x}$&$\alpha$&$\alpha \nabla \alpha$&\textbf{MaRC}&\textbf{ISR}\\
        \midrule
        \textbf{NormSuff} $\uparrow$ & 0.25 & 0.31& 0.28 & 0.26& 0.30 & 0.26& 0.31 & \textbf{0.61}& 0.28 \\
        \textbf{NormComp} $\uparrow$ & 0.45 & 0.46& 0.51 & 0.43& 0.45 & 0.42& 0.53 & \textbf{0.78}& 0.62 \\
        \midrule
        \textbf{F1-Suff} $\uparrow$  & 0.51 & 0.56& 0.68 & 0.54& 0.57 & \textbf{0.73}& 0.57 & 0.58& 0.60 \\
        \textbf{F1-Comp} $\downarrow$& 0.54 & \textbf{0.41}& 0.55 & 0.52& 0.48 & 0.44& 0.54 & 0.44& 0.52 \\

        \bottomrule
    \end{tabular}
    \caption{Quantitative analysis results for the interpretability techniques (recall that ISR is an ensemble technique combining $\alpha$, $\alpha \nabla \alpha$, IG, $\mathbf{x} \nabla \mathbf{x}$, Random, LIME, and DeepLift), with random attribution serving as the baseline. The first two rows report faithfulness metrics, while the last two rows present performance metrics, measured relative to the model’s performance on the original input (0.77 F1-score).}
    \label{tab:suff_comp}
\end{table*}
\section{Results} \label{sec: results}
\subsection{Downstream classification results} \label{6_1_classification_results}

We fine-tune LEGAL-BERT on the ECtHR cases and obtain an F1 score of 0.77. Because the Article~6 and Article~8 subsets do not contain enough data for separate fine-tuning, we evaluate this general model on the article-specific portions of the test set, achieving F1 scores of 0.76 and 0.83 for Article~6 and~8, respectively. Although roughly one third of the dataset concerns Article~6, half of the mispredicted cases fall into this subset. By contrast, Article~8 cases (about 20\% of the dataset) account for only 10\% of the mispredictions. We hypothesize that the procedural complexity of Article~6 makes violation detection more difficult for the model compared to the clearer substantive rights protected by Article~8.

\subsection{Quantiative analysis results of rationale extraction techniques} \label{6_2_quant_results}
We use each feature-attribution method employed by ISR as a comparative reference for faithfulness evaluation of both ISR and MaRC (Table~\ref{tab:suff_comp}). We also use random attribution as a baseline to represent chance-level performance across all metrics.

\paragraph{Sufficiency evaluation} Our experiments indicate that MaRC extracts the most sufficient rationales among all evaluated techniques. By contrast, ISR and the other methods obtain values that are close to the baseline, which is expected given ISR's ensemble architecture. Notably, instance-specific feature attribution does not guarantee optimal performance: ISR would have achieved higher sufficiency scores if it had relied only on LIME or scaled-attention. 

\paragraph{Comprehensiveness analysis}
In contrast to sufficiency assessment, ISR performs better than most techniques, although MaRC still ranks highest. Remarkably, choosing the best technique per instance has a positive impact on comprehensiveness, unlike its effect on sufficiency.

\paragraph{Performance analysis}
Rationale impact on classifier performance does not align with faithfulness assessment. Despite its poor NormSuff results, $\alpha$ achieves the best F1-Suff score, and LIME attains the best F1-Comp score even though it is not among the most faithful techniques. This depicts utility metrics are insufficient for rationale evaluation, as they use classified labels, whereas faithfulness metrics rely on the classifier's confidence.

Our quantitative analysis also shows that sufficiency and comprehensiveness should not be assessed in isolation: considering only one of these metrics risks incomplete or misleading conclusions about rationale quality.

\subsection{Legal expert analysis} 

Models often predict violations on grounds that the legal expert finds unsatisfactory. Expert~B evaluated whether rationales are both supportive and sufficient, and found that they usually satisfy neither criterion. For ISR, nine out of ten documents fail both; for MaRC, seven out of ten do. Thus, Expert~B generally judges the rationales unconvincing and would, in most cases, require additional information to conclude that a violation occurred.

The key reason is that the rationales are often not comprehensible. From Expert~B's perspective, many resemble broken snippets. The techniques frequently extract random nouns without verbs (e.g., \texttt{`non-enforcement (section 74)'}), fragments without subjects (e.g., \texttt{`stated that, based'}), irrelevant dates (e.g., \texttt{`000.8. on 28'}), or interrupted clauses without meaning (e.g.,~\texttt{`. 26/77, 28/77 , 43/77 , 20/79, 24'}). Expert~B also noted that rationales sometimes omit critical terms immediately before or after the highlighted span, so key legal facts are cut off mid-sentence and stripped of their meaning. Such deficiencies render the rationales both insufficient and unsupportive (Figure~\ref{fig: rat_snippet}).
\begin{table*}[h]
\resizebox{\textwidth}{!}{%
\begin{tabular}{l|rrr|rrr}
\toprule
 & \multicolumn{3}{c|}{Sufficiency} & \multicolumn{3}{c}{Support} \\
\midrule
Model & Single-shot & Few-shot & Few-shot (formal) & Single-shot & Few-shot & Few-shot (formal) \\
\midrule
Gemma   & -0.03 [-0.35, 0.31] & 0.20 [-0.15, 0.53] & 0.21 [-0.13, 0.53] & -0.23 [-0.55, 0.10] & \textbf{0.15} [-0.13, 0.42] & 0.05 [-0.20, 0.30] \\
Llama   & \textbf{0.40} [0.05, 0.71] & \textbf{0.30} [0.00, 0.60] & \textbf{0.27} [0.02, 0.55] & \textbf{0.29} [-0.04, 0.60] & 0.12 [-0.10, 0.35] & 0.09 [0.00, 0.25] \\
Mixtral & 0.22 [-0.07, 0.53] & 0.20 [-0.14, 0.55] & 0.19 [-0.16, 0.53] & 0.03 [-0.18, 0.29] & 0.13 [-0.13, 0.41] & 0.03 [-0.29, 0.37] \\
\bottomrule
\end{tabular}%
}
\caption{Cohen's $\kappa$ with 95\% CI on global rationale sufficiency criterion. Scores represent agreement of each LLM with Expert B.}
\label{tab:kappa-suff}
\end{table*}
\begin{figure}[h!]
    \small{
         [R1] the case 4 . the; 
         
         [R2] moscow . the first applicant; 
         
         [R3] s father . a . transactions with; 
         
         [R4] the city . 6 . on an; 
         
         [R5] . f . and the; 
         
         [R6] agreement . 7 . on; 
         
         [R7] . f . sold the; 
         
         [R8] to l . m . , the; 
         
         [R9] . m . paid rub; 
         
         [R10] . for the flat .; 
         
         [R11] 000 . 8 . on 28; 
         
         [R12] . m . died . the first; 
         
         [R13] in moscow . 9 . on 10; 
         
         [R14] gift . b . annulment; 
         
         [R15] . m . and that; 
         
         [R16] . f . and the; 
         
         [R17] the neighbours . 11 . on 5; 
         
         [R18] family . 12 . on 15; 
         
         [R19] [UNK] eviction . 13 . on 29; 
         
         [R20] . [UNK] s interest . 14 . on 12; 
         
         [R21] on appeal . 15 . on 13; 
         
         [R22] cassation appeal . c . eviction proceedings 16 .; 
         
         [R23] . as the beneficiary . 17 . on 3; 
         
         [R24] enforcement proceedings . 18 . on 30; 
         
         [R25] . 19 . on 16
    }
    \caption{Rationales extracted by ISR from the case of \href{https://hudoc.echr.coe.int/eng?i=001-175482}{Malayevy v. Russia}.}
    \label{fig: rat_snippet}
\end{figure}

Moreover, human and model rationales differ significantly. In addition to model-based rationales, Expert~A extracted human rationales independently for comparison. First, Expert~A selects whole key paragraphs or uninterrupted substantial parts, providing context that model rationales often lack. Second, Expert~A produces fewer but longer rationales (91 human vs. 108 ISR and 168 MaRC); nevertheless, human rationales are still shorter overall than MaRC's (3,959 vs. 5,183 total tokens) while being far more sufficient. Third, human rationales preserve coherent chronological account of the facts and procedural history—particularly important for right-to-a-fair-trial cases under Article~6—whereas model rationales are much more fragmented.

\subsection{LLM-as-a-Judge} 
We use LLM-as-a-Judge evaluations of rationales to measure how well LLMs align with Expert B. In the few-shot setting, we provide Expert B’s explanations in two forms: their original brief comments and a polished version (the columns labeled formal in Table \ref{tab:kappa-suff}). This allows us to test how the wording of the explanations affects the evaluation (see Appendix \ref{sec: app_prompts}).
\paragraph{Agreement of LLMs with experts}
Table \ref{tab:kappa-suff} reports each LLM’s agreement with Expert B on global rationale sufficiency and support. While Gemma performs very poorly on sufficiency in the single-shot setup, Llama and Mixtral achieve fair agreement with Expert B ($(\kappa \in [0.2, 0.4]$) is typically considered fair). Though, when assessing sufficiency in the few-shot configuration, almost all scenarios reach fair agreement.




However, bootstrapped confidence intervals (CI) indicate that the agreement scores are not reliable. Two factors drive this result. First, the sample of 30 documents offers only directional insight and limited statistical power. Second, LLMs do not assess consistently: while Expert B’s sufficiency and support judgments are stable across rationales, LLMs often disagree on the same document. These findings show that LLM-as-a-Judge results must be interpreted cautiously given hallucination risks and should not be taken at face value.

We also investigate inter-LLM agreement based on these evaluations; detailed results are provided in Appendix~\ref{sec: app_prompts}.

\subsection{Ablation studies}


We also run ablation studies to remove rationale length as a confounding factor. Since each rationale extraction technique has its own (uncontrolled) length, normalized faithfulness metrics can be affected. To make comparisons more objective, we use the RemOve And Retrain (ROAR) \cite{hooker_benchmark_2019} technique (see Appendix \ref{sec: app_config}). Under ROAR, more faithful techniques cause a sharper performance drop when only a small fraction of tokens is removed. Figure~\ref{fig: methods_roar} in the appendix shows that $\alpha$ performs best; IG also drops heavily followed by a spike. MaRC and ISR both drop substantially, with ISR outperforming MaRC, but neither is the top method overall.

We hypothesize that ISR and MaRC depend more on span-level rationales than on local token-level importance (as in LIME). This likely explains why they perform best under normalized evaluation but only moderately under ROAR. Overall, these results indicate inconsistencies between automatic evaluation of rationale extraction techniques because NormSuff, NormComp, and ROAR disagree on whether ISR or MaRC scores better.


\section{Conclusion} \label{sec: conclusion}



We propose a standardized comparative framework to evaluate rationale extraction techniques for legal outcome prediction on a new ECtHR dataset. By combining faithfulness metrics with legal expert analysis, we provide a more holistic view of rationale quality and also test the feasibility of LLM-as-a-Judge for plausibility evaluation, aiming to mitigate the scalability limits of expert assessment.
We show that model-agnostic extractions diverge sharply from the reasons legal professionals rely on. This lack of genuine legal grounds for model conclusions makes current systems problematic for real-world use, where decisions must be properly justified. We further demonstrate that contemporary open-weights LLMs are not yet reliable enough for LLM-as-a-Judge in the legal domain.


\section{Ethical considerations} \label{sec: ethics}
In this work we use publicly available ECtHR cases. Our comparative analysis aims to improve the transparency in legal NLP applications without any aim to replace legal expert judgment.

\section{Limitations}
Our work is limited to English language data, despite very few details were explained in the case documents in various languages. Furthermore, our comparative evaluation is limited with model-agnostic interpretability techniques, because being inherently dependent on the task and architecture hinders us to employ task-specific interpretability  methods.
The number of only ten manually analyzed documents might seem small, but the overall costs of the expert annotation study were a hard limiting factor (thousands of EUR).
We also relied on open-weight LLMs and did not use proprietary models to ensure reproducibility, transparency, and to prevent future data contamination. Note that we consider this decision as a research design choice, not as an inherent limitation.

\bibliography{references}

\appendix

\section{Configuration details} \label{sec: app_config}
\subsection{Classifier} We use LEGAL-BERT \cite{chalkidis-etal-2020-legal} for the downstream classification task of predicting violation exitence on ECtHR cases, due to the model's legal pre-training. This choice comes with a limitation: LEGAL-BERT has a context window of 512 tokens. We address this by combining the data processing strategy from MaRC \cite{brinner_model_2023} with the hierarchical configuration of HIER-BERT \cite{chalkidis_extreme_2019}. The former splits long documents into chunks, and the latter processes these chunks using a cross-attention mechanism. Unlike MaRC, we do not sample a single chunk to represent the whole document; instead we process all chunks before performing classification.

For the classification task, we use the subsampled subset of our new ECtHR dataset (Table \ref{tab: subset_statistics}), which is constructed specifically for this study (Section~\ref{sec: downstream}).

\begin{table}[h!]
    \centering
    \begin{tabular}{ll|rrr}
        \toprule
        \textbf{Dataset} & \textbf{Violation}&\textbf{Train} & \textbf{Valid.} & \textbf{Test}\\
        \midrule
        \multirow{2}{*}{Subset}&Positive & 1091 & 315 & 168\\
        &Negative & 1091 & 315 & 166\\
        \midrule
        \multirow{2}{*}{Article 6}&Positive & 399 & 116& 64\\
        &Negative & 400 & 116& 63\\
        \midrule
        \multirow{2}{*}{Article 8}&Positive & 228 & 66 & 39\\
        &Negative & 228 & 67 & 38 \\
        \bottomrule
    \end{tabular}
    \caption{Distribution of positive (violation) and negative (no violation) cases in the subsets of the new ECtHR dataset that is used for this study. Article 6 and Article 8 correspond to the subsets of the dataset that involves corresponding articles.}
    \label{tab: subset_statistics}
\end{table}
\subsection{MaRC}
\subsubsection{Mathematical background}
The interpretability method extracts class-indicative text fragments from the given input text \textit{x}, by optimizing the mask $\lambda$. \citet{brinner_model_2023} define a continuous mask vector $\lambda\in[0,1]^n$, one value per token. Each value of this vector makes the masked input token ($\Tilde{x}_i$) interpolates between the original input ($x_i$) and uninformative input ($b_i$) token (Equation \ref{eq: masked_input}).
\begin{equation} \label{eq: masked_input}
    \Tilde{x} = \lambda \cdot x + (1-\lambda) \cdot b
\end{equation}

Mask is optimized to satisfy sufficiency, comprehensiveness and compactness constraints were introduced by \cite{yu_rethinking_2019}. Each constraint is handled by MaRC by constraint-specific regularizer. Notice that, mask is used to obtain the rationales, rather than excluding non-informative fragments from the text input.
\paragraph{Sufficiency} For a given class $c$, MaRC minimizes the difference between the classifier’s output on the original input and on the masked input, so that the masked input is sufficient to replace the full text. To achieve this, the authors propose the following optimization objective:
\begin{equation} \label{eq: sufficiency_opt}
    \underset{{\lambda \in [0, 1]^n}}{\arg\min} -\mathcal{L}(\Tilde{x}, c) + \underbrace{\alpha_\lambda\left[\frac{1}{n}\sum_{i=1}^n{\lambda_i}\right]^2}_{\Omega_\lambda}
\end{equation}

$\mathcal{L}(\tilde{x}, c)$ denotes the log-likelihood of class $c$ under the classifier model $M$, which scores the input tokens to maximize the probability of $c$. The second term in the equation is a sparsity regularizer $\Omega_\lambda$ that encourages the set of high-scoring tokens for $c$ to be as small as possible.

\paragraph{Comprehensiveness} To specify the significance of the class-indicative information, authors penalize classifier's confidence on the complementary masked input ($\Tilde{x}^\texttt{c}$) as
\begin{equation} \label{eq: complement}
    \Tilde{x}^\texttt{c} = (1 - \lambda)\cdot x + \lambda\cdot b
\end{equation}
To enforce comprehensiveness in the rationales, \citet{brinner_model_2023} update the given objective function above by adding scoring function for such tokens:
\begin{equation} \label{eq: sufficiency_opt}
    \underset{{\lambda \in [0, 1]^n}}{\arg\min} -\mathcal{L}(\Tilde{x}, c) + \mathcal{L}(\Tilde{x}^\texttt{c}, c) +{\Omega_\lambda}
\end{equation}
\paragraph{Compactness} This term encourages the selection of connected spans rather than isolated tokens. It is enforced by reparameterizing $\lambda$, so that the mask is not optimized directly. The authors define the mask using two new parameters, $\omega \in \mathbb{R}^n$ (token weights) and $\sigma \in \mathbb{R}{>0}^n$ (the extent to which each token influences its neighbors). The impact of the $i^{\text{th}}$ token’s weight on the weight of the $j^{\text{th}}$ token is computed as
\begin{equation} \label{eq: weight_impact} 
w{i \rightarrow j} = w_i \cdot \exp\left(-\frac{d(i,j)^2}{\sigma_i}\right), \end{equation}
where $d(i,j)$ denotes the distance between the $i^{\text{th}}$ and $j^{\text{th}}$ tokens. The mask value is then computed as \begin{equation} \label{eq: lambda_comp} 
\lambda_j = \text{sigmoid}\left(\sum_i w_{i \rightarrow j}\right). \end{equation}
Because contiguity is important for rationales, large values of $\sigma_i$ are crucial in this setup. As shown in Equation~\ref{eq: weight_impact}, larger $\sigma_i$ values cause $w_i$ to spread its influence over neighboring tokens. Such large $\sigma_i$ values are encouraged by the compactness regularizer: \begin{equation} \label{eq: compactness} 
\Omega_\sigma = -\alpha_\sigma \cdot \frac{1}{n} \sum_{i=1}^{n} \log(\sigma_i)
\end{equation}
The final MaRC objective combines this term with the sufficiency and comprehensiveness terms: 
\begin{equation} \label{main_objective}
\underset{w, , \sigma \in \mathbb{R}^n}{\arg\min} -\mathcal{L}(\tilde{x}, \mathsf{c}) + \mathcal{L}(\tilde{x}^c, c) + \Omega_\lambda + \Omega_\sigma
\end{equation}

\subsubsection{Implementation details}
We do not change the hyperparameters or configuration of the technique unless necessary. We set $\alpha_{\lambda}$ and $\alpha_{\sigma}$ to 1 and 1.2, respectively, and initialize $\omega$ and $\sigma$ uniformly to 1.2 and 2. In addition, we add zero-mean Gaussian noise with standard deviation 0.3 to the masked input ($\tilde{x}$) and its complement ($\tilde{x}^{\mathsf{c}}$). Finally, 5\% of mask values are set to 0 and 1 at random for the masked input and its complement, respectively.
\subsection{Flexible Instance Specific Rationale Extraction} We modify the configuration of the ISR to make it applicable in legal domain. In its original form, ISR aims to extract a single best combination out of all possible combinations of $N$ different lengths, two rationale types (\textbf{top-K} and \textbf{Contiguous-K}) and seven feature-attribution methods. We instead configure ISR to extract multiple rationales from the input text, since a single rationale is not sufficient to decide a legal outcome in the real-world applications. 

Because the input texts are much longer than in the SST dataset on which ISR was originally evaluated \cite{chrysostomou_flexible_2021}, we process them in batches, so that feature-attribution is applied per input text. In our configuration, we create a dynamic-size batches, where the size of the batch is determined by the number of chunks obtained via chunking mechanism of MaRC \cite{brinner_model_2023}. ISR assign significance scores to these chunks before classification, since prediction is made at the case level rather than separately for each chunk.

Furthermore, we restrict the rationale type to \textbf{contiguous-K}, since selecting sparse tokens from distant parts of the input is not an optimal way to justify decisions in legal applications. We therefore collect all adjacent token sequences of predefined length $l \in [1, \dots, N]$ with a window size of 1. Each candidate sequence is scored by summing its token-level scores, and overlapping sequences are merged by averaging their scores. In this way, the technique adapts the rationale length to the specific case. For each $l$, we retain sequences whose scores exceed the average score of all combinations for that $l$ as candidate rationales. Finally, these candidates are combined with the different importance-scoring methods to compute the divergence.

Finally, we do not re-evaluate the divergence scoring techniques used by \citet{chrysostomou_flexible_2021}. Instead, we adopt the divergence measure that achieved the best results in their work, Jensen–Shannon divergence (JSD) \cite{lin1991divergence}, as the most suitable choice for our setting.
\paragraph{Scoring techniques in ISR}
ISR employs seven different scoring techniques for rationale extraction:
\begin{itemize}
    \item Random: This method assigns token importance score randomly; 
    \item Attention ($\alpha$): Normalized attention scores are used to assign token importance \cite{jain_learning_2020}
    \item Scaled attention ($\alpha\nabla\alpha$): The method assigns scaled attention scores by their gradients \cite{serrano_is_2019}
    \item Integrated gradient: Scores are assigned by the integral of the gradients, which is computed along the path from zero embedding vector to the original input \cite{sundararajan_axiomatic_2017};
    \item InputXGrad ($\mathbf{x}\nabla\mathbf{x}$): The method computes the significance scores by scaling each element of the input with the gradient of the model's outcome with respect to the input \cite{kindermans_investigating_2016, atanasova_diagnostic_2020}
    \item DeepLift: Method compares the outcome of the actual and baseline inputs (e.g., zero embedding vector) by computing the changes in the activations of neurons to assign significance scores to the features \cite{shrikumar_learning_2017}.
    \item LIME: Scores are given by approximating model's behavior locally with an interpretable model \cite{ribeiro_why_2016}.
\end{itemize}
\subsection{RemOve And Retrain (ROAR)} ROAR \cite{hooker_benchmark_2019} is a model-agnostic evaluation metric that assesses the quality of rationales. The evaluation process consists of the following steps:
\begin{itemize} 
    \item \textbf{Identify important features:} Using feature attribution methods, the individual methods that make up ISR’s ensemble, ISR itself, and MaRC, we score all features in the input texts for the training, validation, and test sets.
    \item \textbf{Remove the features:} After scoring, we rank features according to their scores and mask them under two configurations: (a) masking the highest top $k\%$ of scores; and (b) masking the \emph{absolute} highest top $k\%$ of scores across the dataset. The first configuration follows the rationale extraction methodology; the second is expected to provide more reliable results, especially when significance scores can also take negative values.
    \item  \textbf{Retrain the classifier:} After masking rationales out of the input, we retrain the model on the resulting data. If the extracted rationales are of high quality, we expect a drop in performance.
\end{itemize}

The technique helps us assess rationale quality while controlling for rationale length, since masking is applied to the same proportion of features for each feature-attribution method. It evaluates quality by tracking the change (ideally, a drop) in classifier performance as $k$ is varied from 0 (original input) to 100 (no informative feature remaining).

We note that our evaluation removes tokens ranked by positive attribution weights. This design choice is intentional, as rationales are defined as tokens that positively support the prediction. However, this means tokens with strong negative weights are not removed in the first 2 or 3 (depending on the range) iterations, which may contribute to the observed robustness of certain methods at high removal percentages. Evaluating the effect of removing tokens by absolute attribution magnitude is left for future work.
\begin{figure*}
    \centering
    \includegraphics[width=1\linewidth]{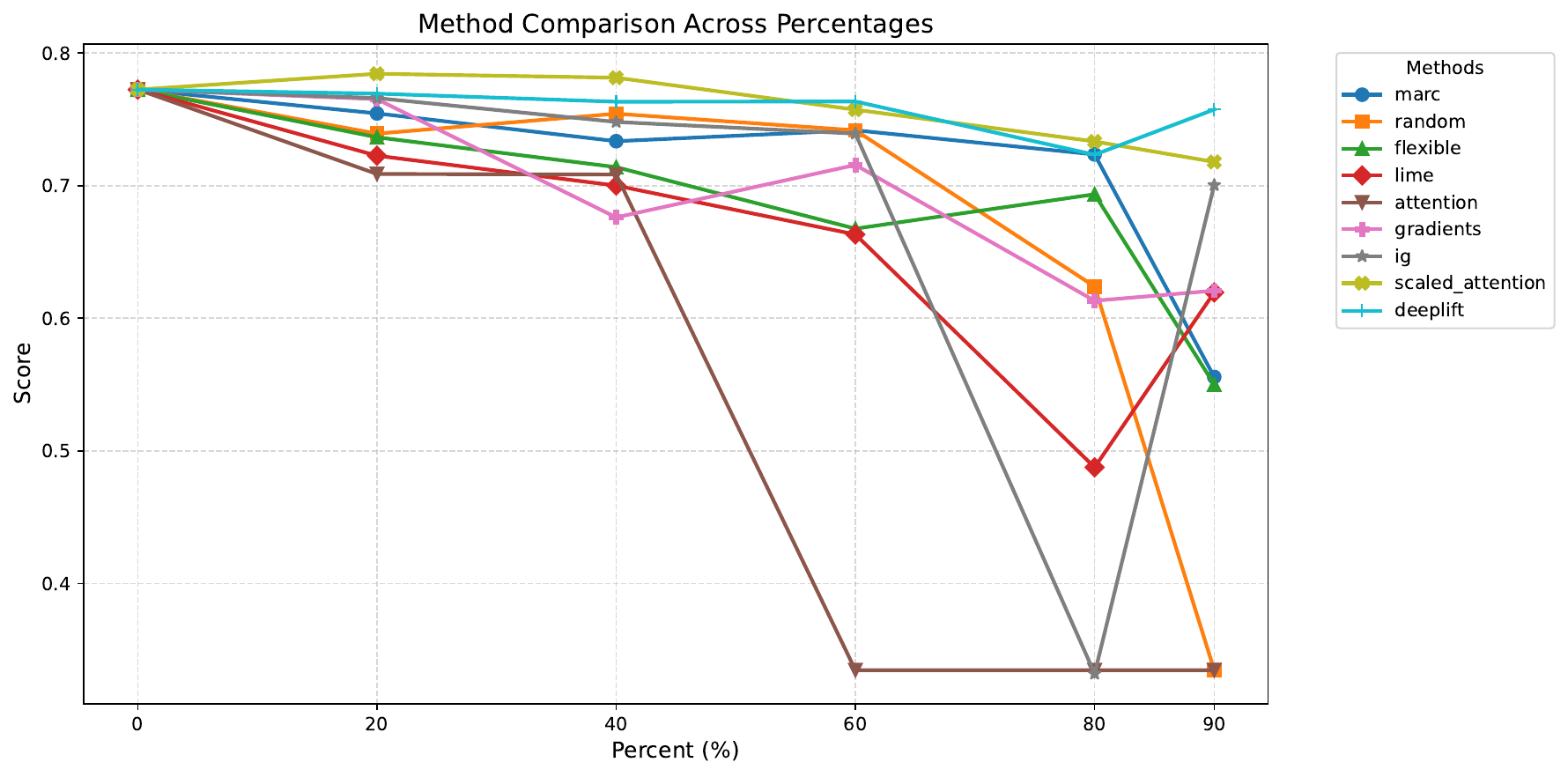}
    \caption{ROAR evaluation results for all the methods.}
    \label{fig: methods_roar}
\end{figure*}

\section{Creating the new ECtHR dataset} \label{sec: app_data}
We create a new dataset, since previous datasets lacked sufficient admissible data for our classification task. To create the new dataset, we collaborate with a legal expert to define filters. We use these filters to distinguish positive (violation) and negative (non-violation) cases. Furthermore, we add few more filters to eliminate ambiguous cases, where violation existence cannot be conclusively determined.

\paragraph{Inadmissible} There are certain criteria that a case must meet to be considered by the ECtHR, which are set out by Article 34 \footnote{Article 34 of the ECHR: Standing: determines who can apply to the ECtHR.} and Article 35 \footnote{Article 35 of the ECHR: Admissibility conditions: exhaustion of domestic remedies, no anonymous application and etc.}. If these requirements are not met, the case is declared inadmissible. In that situation, the application is rejected and the Court does not examine the case on its merits. Because the Court does not assess the merits, it is not possible to say whethere there is a violation in cases it deems inadmissible. It should be noted that around 90\% of applications to the ECtHR are considered inadmissible. For this reason, we do not include inadmissible cases in our dataset. Although this reduces the dataset size, it makes our classification task and its results more reliable.

\paragraph{Violation} The term refers to a breach of a legal norm as resulting from an action or omission. Following \citet{chalkidis_neural_2019}, we consider a case as positive if any article is found to be violated. Accordingly, we use the word "Violation" as one of the keywords to detect cases with violations. Note that this is not the only filter we use to identify cases in which an article is violated.

\paragraph{Award} The word refers to the measures taken to restore or improve the situation of the victim in the event of a violation. If the word \emph{award} appears in the conclusion of the case (e.g., “Non-pecuniary damage - financial award”, “Costs and expenses partial award - Convention proceedings”), then there is a violation and an award is also determined. Therefore, we keep such cases in the dataset as positive samples, using this word as one of our filter.

\paragraph{Finding of violation sufficient} The phrase itself indicates that the fact of a violation has been confirmed by the Court. Therefore, the presence of this expression helps us to identify positive cases.

\paragraph{Struck out of the list} According to Article 37\footnote{Article 37 of the ECHR: Striking out applications} of the Convention, the Court may remove a case from ist list for various reasons (for example, at the applicant's request, the issue has been resolved, or due to any event that makes further consideration unnecessary). In such situations, the merits of the case are not assessed, except where the Court later decides that the issue should be reconsidered. Therefore, we cannot determine whether there is a violation or not, and we do not include such cases in the dataset.

\paragraph{Lack of jurisdiction} When the Court cannot consider a case, it may declare a \emph{lack of jurisdiction} (e.g., the complaint does not concern rights protected by the Convention and its Protocols, or when the alleged violation is not attributable to the State). In such cases, we cannot determine the existence of any violation. Thus, we do not include these cases in our dataset.

\paragraph{Not necessary to examine} When the conclusion includes this phrase bound to an article, we do not apply any filtering to that article. However, if the Court’s conclusion contains other information (for example, a clear finding of a violation for another article), we still apply our filters to those remaining articles to detect possible violations.

\paragraph{Government's request to strike the application out of the list rejected} In some cases, the government (i.e., the respondent State) requests that the Court refrain from examining the case on its merits. However, if the Court decides that the case should be considered, it rejects this request and examines the case on its merits. In such cases we discard government's request and apply filter to the rest of the conclusion.

\paragraph{Preliminary objection allowed (or Preliminary objection partially allowed)} Preliminary objection is raised by the government in a given case, typically invoking inadmissibility or lack of jurisdiction. If the Court accepts the objection in its entirety, no investigation is conducted on the merits. However, if the Court agrees with the government only in part, the case is investigated only partially. Because the existence of a violation is ambiguous in such situations, we do not include these cases in our dataset.

\section{Legal expert evaluation} \label{sec: app_ratex} 
Expert A follows evaluation guidelines of \citet{chalkidis_paragraph-level_2021}, who similarly employ a legal expert to annotate the rationales from the ECtHR cases in the similar manner. For expert B, we develop our guidelines as there are not task-specific instructions in the literature. These guidelines are broader than what we use in this paper and also cover local rationale support and ex-post sufficiency. In our experiments, however, we found that neither local rationale support nor ex-post sufficiency provided sufficiently informative signals for the legal application. Thus, we focus here on the global support and sufficiency criteria.
\subsection{Annotation guidelines}
\subsubsection{General descriptions} 
\paragraph{Annotation process} Annotators are asked to follow the steps below when performing the annotations:
\begin{itemize}
    \item Read the violated article(s) in the case concerned.
    \item Evaluate only global rationale support and global rationale sufficiency criteria. Provide also your confidence level to the answer with one of the letters of L, M and H, which stand for low, medium and high confidence, respectively;
    \item Evaluate the local rationale support criterion after the previous 2 criteria analysis is done;
    \item Evaluate post-sufficiency analysis as in the sufficiency analysis. This time, you may draw on your broader knowledge of the legal case.
\end{itemize}
\paragraph{What is a rationale?}
An uninterrupted area of highlighted text counts as a single rationale. Even if the rationale is interrupted only by a single non-highlighted word, it splits one rationale into two. These two rationales then receive distinct labels. Even when the highlighted area of text crosses from one paragraph to the next, we still mark it as a single rationale and assign one label.
\subsubsection{Evaluation criteria}
Guidelines for each criterion were tailored through several iterations.
\paragraph{Global rationale support} 
\begin{itemize}
    \item Annotate each document with either "Support" or "Unsupport";
    \item Annotate as "Support", if you think that all rationales together support the outcome (i.e., the existence of a violation); otherwise, use "Unsupport".
    \item Do not consider whether you have all the information needed. Instead, focus only on  whether the given rationales support the outcome.
    \item Assign also confidence level, as explained above.
\end{itemize}

\paragraph{Global rationale sufficiency}
\begin{itemize}
    \item Assign a binary value of either “Sufficient” or “Insufficient” per case text.
    \item Consider whether you have more or less all the necessary information needed to infer that the alleged violation really occurred. If you require more information, annotate “Insufficient”. 
    \item Assign also a confidence level, as explained above;
\end{itemize}
\paragraph{Local rationale support}
\begin{itemize}
    \item Assign a value of either “Support” or “Unsupport” per rationale. If you choose "Unsupport", select one of four subcategories:
    \begin{itemize}
        \item Relevant but unsupportive: The rationale is relevant to the conclusion but not does not support it.
        \item No violation: The rationale supports contrary conclusion.
        \item Irrelevant: The rationale is irrelevant to the conclusion.
        \item Incomplete/conflict: The rationale does not make sense as a whole, or one part supports the outcome, while another part points in the opposite direction.
    \end{itemize}
    \item “Supports” does not mean that the rationale sufficiently supports the conclusion. This category has a lower threshold: anything that provides some support for the conclusion counts as "Support".
    \item There is no confidence score for these annotations.
    \item Use the case context mainly to understand the story behind the case, what happened, and what the alleged violation is, so that you can assess whether the particular rationale supports the conclusion about a violation. Do not use the context to infer what the model should have highlighted bu did not.
\end{itemize}
\paragraph{Ex-post sufficiency}
\begin{itemize}
    \item Read the entire decision.
    \item Review all model-provided rationales.
    \item Answer the following question: “Given these rationales and the context of the case and the violated articles, can I confidently say that a violation really occurred based on the rationales alone?”
    \item Reflect on whether the model has missed information that is necessary to conclude that a violation occurred. Use the full context to decide whether such information is missing.
    \item If the rationales are sufficient, mark "Sufficient"; otherwise, mark Insufficient.
\end{itemize}
\paragraph{Special circumstances} As a default rule, annotators should take the highlighted texts at face value. Annotators must not infer beyond the highlighted span unless special circumstances apply.
\begin{itemize}
    \item Linguistic phenomena within the rationale:
    \begin{itemize}
        \item If the rationale is a sentence with an unstated subject, you may look for the subject in the surrounding text.
        \item If the highlighted text contains a pronoun, you may look for its antecedent.
        \item If the rationale uses a definition given elsewhere, you may consult that definition.      
    \end{itemize}
    \item If the highlighted text contains “the heart of the matter”, even when it is surrounded by random or neutral context, you may treat that core content as carrying the main meaning.
    \item When the AI model omits important context of the sentence/paragraph:
        \begin{itemize}
            \item Label a rationale as “Unsupport” when the model omits context that would change the rationale from supporting to not supporting the conclusion. If the complete sentence would contradict or invalidate the supportive impression given by the rationale alone, label it as “Unsupport".
            \item If the model omits context that cannot reasonably be inferred in the rationale’s favor and that context would be necessary to label the rationale as “Supports”, then label it as “Unsupport”.
        \end{itemize}

\end{itemize}
\begin{table}[t]
    \centering
    \setlength{\tabcolsep}{4pt}
    \begin{tabular}{l|rr}
        \toprule
        \textbf{Metric} & \textbf{MaRC} & \textbf{ISR}  \\
        \midrule
        Number of rationales & 9,828 & 15,149 \\
        Average length of rationale & 17.63 & 12.57 \\
        Number of rationales per doc & 29.42 & 45.36\\
        \bottomrule
    \end{tabular}
    \caption{Rationale statistics for each rationale extraction technique}
    \label{tab:rat_stats}
\end{table}

\subsection{Training phase} 
After the pilot studies, all experts assessed ten documents (i.e., five documents processed by each technique). Each annotator spent roughly one hour per decision, and inter‑annotator agreement was tracked automatically in INCEpTION \cite{tubiblio106270}. A majority vote determined the final label set for local rationales (Support/Unsupport). Inter‑coder agreement statistics are reported in Table \ref{tab: local_rat}. For this computation we employ Krippendorff's $\alpha$ \cite{krippendorff2018content}, as it is suitable for more than two coders. We emphasize that these five documents are different from those evaluated in the main study.

After careful evaluation, we decided to discard the local rationale evaluation from this study. Our expert analysis shows that single rationales convey very little information about the outcome, whereas the set of rationales taken together can still be informative.
\begin{table}[h]
    \centering
    \begin{tabular}{l|rr}
        \toprule
        \textbf{Technique} & \textbf{MaRC} &\textbf{ISR} \\
        \midrule
        Num. rationales & 159 & 170 \\
        Support/Unsupport &  13/146 & 8/162 \\
        Krippendorff's $\alpha$ & 0.73& 0.53 \\
        \bottomrule
    \end{tabular}
    \caption{Local rationale support evaluation results on the chosen five ECtHR cases, were evaluated on the previous version of this study}
    \label{tab: local_rat}
\end{table}
\section{Extracted rationales} The Figures \ref{fig: cont_385} and \ref{fig: cont_582} depict examples of rationales within the full context of the legal case. We observe that, in most cases, rationales by MaRC are longer than those extracted by ISR (see Table \ref{tab:rat_stats}). However, the number of rationales produced by ISR is higher than by MaRC. The lower F1-Comp value for MaRC compared to ISR can be explained by the length of the rationales, since we mask this information when computing F1-Comp. Our legal expert analysis, however, shows that much of the information conveyed by these rationales is only weakly relevant to the outcome. Thus, it is not clear to what extent the classifier actually captures the underlying legal concepts.

         
         
         
         
         
         
         
         
         
         
         
         
         
         
         
         
         
         
         
         
         
         
         
         

For global rationale support and sufficiency criteria analysis, we provide the legal experts with a version of each document that contains only the highlighted rationales, as illustrated in Figures ~\ref{fig: cont_385_rat} and~\ref{fig: cont_582_rat}. 

As discussed in the legal expert analysis section of the paper (Section~\ref{subsec: qualysis}), some rationales do not carry significant information for assessment. Figure~\ref{fig: rat_snippet} shows all rationales extracted from the case of Malayevy v. Russia, which are uninformative. We then scrutinize LLM-as-a-Judge evaluation of these rationales in Figures \ref{fig: support_malayevy} and \ref{fig: sufficiency_malayevy}.

\begin{figure*}
    \begin{mdframed}[linewidth=.5pt, roundcorner=10pt, backgroundcolor=gray!1]
        \begin{subfigure}{1\textwidth}
            \small{
                district court once again ordered their return to spain . the decision was upheld in the subsequent procedure by the high court and the \uline{kuria . 6 . all of these instances relied on evidence} such as the common apartment of the family until the separation , the financial contribution of the parties to the family and the registration of the children in local communities , nurseries and health 
                \uline{care services . 7 . on 21 october 2019 the constitutional court suspended anew the execution of the return proceedings and , } on 25 february 2020 , cancelled
            }
            \caption{Rationales (with context) by MaRC}
            \label{subfig: cont_385_1}
        \end{subfigure}%
        \vspace{0.6cm}
        \begin{subfigure}{1\textwidth}
            \small{
                ... these instances relied on evidence such as the common apartment of the family until the separation , the financial contribution of the parties to the family \uline{and the registration of the children in local communities , nurseries and health care services . 7 . on 21 october 2019 the constitutional court suspended anew the execution of the return proceedings and , on 25 february 2020 , cancelled} the last procedure on the grounds that the psychological impact of the return to spain on the children had been insufficiently evaluated and the mother had not had the opportunity ...
            }
            \caption{Rationales (with context) by ISR}
            \label{subfig: cont_385_2}
        \end{subfigure}
        \vspace{0.2cm}
        
        \begin{subfigure}{1\textwidth}
            \small{
                ... \uline{On 25 March 2019, having obtained a report on the psychological evaluation of the children, the Pest Central District Court once again ordered their return to Spain. The decision was upheld in the subsequent procedure by the High Court and the Kúria.} 
                6. All of these instances relied on evidence such as the common apartment of the family until the separation, ...
            }
            \caption{Rationales (with context by expert A)}
            \label{subfig: cont_385_3}
        \end{subfigure}
        
    \end{mdframed}
    \caption{Extracted rationales by MaRC (a), ISR (b) and expert A (c), from the same span of the case of \href{https://hudoc.echr.coe.int/eng?i=001-228385}{Vassallo v. Hungary, 2023}. As mentioned in the legal analysis part, MaRC extracts longer rationales than ISR, usually. However, it does not always mean that it carries supportive information for the outcome.}
    \label{fig: cont_385}
\end{figure*}

\begin{figure*}
    \begin{mdframed}[linewidth=.5pt, roundcorner=10pt, backgroundcolor=gray!1]
        \begin{subfigure}{1\textwidth}
            \small{
                    1 . the case concerns the alleged failure of the hungarian authorities to conduct a swift examination in proceedings under the hague convention on the civil aspects of international child abduction ( [UNK] hague convention [UNK] ) . 2 . in 2009 the applicant met ms b . , a hungarian; ibiza alone . b . went to hungary; february 2017 the applicant filed an application for the return of; and 2018 the budapest high court and subsequently the kuria confirmed this decision . 4 . on 13 february 2018 the constitutional court suspended the enforcement of the return orders and on 27 november 2018 cancelled the previous decisions on the grounds that
             }
            \caption{Concatenated rationales, which were extracted by MaRC}
            \label{subfig: cont_385_rat_1}
        \end{subfigure}%
        \vspace{0.3cm}
        
        \begin{subfigure}{1\textwidth}
            \small{
                1 . the case concerns the alleged failure of the hungarian authorities to conduct a swift examination in proceedings under the hague convention on the civil aspects of international child abduction ( [UNK] hague convention [UNK] ); and the registration of the children in local communities , nurseries and health care services . 7 . on 21 october 2019 the constitutional court suspended anew the execution of the return proceedings and , on 25 february 2020 , cancelled; where they have remained with him since , according to the elements in the case file . 12 . on 24 november 2020 the
            }
        \caption{Concatenated rationales, which were extracted by ISR}
        \label{subfig: cont_385_rat_2}
        \end{subfigure}
        \vspace{0.1cm}
    
        \begin{subfigure}{1\textwidth}
            \small{            
                In 2009 the applicant met Ms B., a Hungarian national. They lived together for several years in Spain, where they were married in 2015. From this union two children were born, in 2013 in Hungary and in 2015 in Spain. In January 2017, after a family holiday with the children in a third country, the applicant returned to their home in Ibiza alone. B. went to Hungary with the children and announced to the applicant her intention to settle there permanently.; On 15 February 2017 the applicant filed an application for the return of the children to Spain based on the Hague Convention. On 13 July 2017, concluding that the children’s habitual residence was in Spain, the Pest Central District Court ordered their return. In 2017 and 2018 the Budapest High Court and subsequently the Kúria confirmed this decision.            
            }
        \caption{Concatenated rationales, which were extracted by expert A}
        \label{subfig: cont_385_rat_3}
        \end{subfigure}

    \end{mdframed}
    \caption{Example from concatenated rationales by MaRC (a), ISR (b) and expert A (c), from the case of \href{https://hudoc.echr.coe.int/eng?i=001-228385}{Vassallo v. Hungary, 2023}. Expert B assesses global support and sufficiency criteria using such documents.}
    \label{fig: cont_385_rat}
\end{figure*}

\begin{figure*}
    \begin{mdframed}[linewidth=.5pt, roundcorner=10pt, backgroundcolor=gray!1]
        \begin{subfigure}{1\textwidth}
            \small{
                \uline{the circumstances of the case 4 . the applicant was born in 1979 and lives in kosice . a . proceedings concerning the purchase of a flat 5 . on 30 march 1999 the applicant filed a civil action with the kosice ii district court . she claimed} that the defendant should be ordered , in accordance with his earlier written undertaking , to sell a flat to her . 6 . the applicant replied to the defendant \uline{on 23 july 1999 and 8 february 2000 . 7 . in the course of the proceedings the applicant discovered} that the defendant had transferred the owner ...
            }
            \caption{Rationales (with context) by MaRC}
            \label{subfig: cont_582_1}
        \end{subfigure}%
        \vspace{0.3cm}
        \begin{subfigure}{1\textwidth}
            \small{
                the circumstances of \uline{the case 4 . the applicant was born in 1979 and lives in kosice . a . proceedings concerning the purchase of a flat 5 . on 30 march 1999 the applicant filed a civil action with the kosice ii district court . she} claimed that the defendant should be ordered , in accordance with his earlier written undertaking , to sell a flat to her ...

            }
            \caption{Rationales (with context) by ISR}
            \label{subfig: cont_582_2}
        \end{subfigure}
        \vspace{0.3cm}
        \begin{subfigure}{1\textwidth}
            \small{ 
                ... The applicant was born in 1979 and lives in Košice. A. Proceedings concerning the purchase of a flat \uline{5. On 30 March 1999 the applicant filed a civil action with the Košice II District Court. She claimed that the defendant should be ordered, in accordance with his earlier written undertaking, to sell a flat to her.} 6. The applicant replied to ...
            
            }
            \caption{Rationales (with context) by expert A}
            \label{subfig: cont_582_3}
            
        \end{subfigure}
        
    \end{mdframed}
    \caption{Extracted rationales by MaRC (a), ISR (b) and expert A (c), from the same span of the case of \href{https://hudoc.echr.coe.int/eng?i=001-92582}{Buľková v. Slovakia, 2017}. As mentioned in the legal analysis part, MaRC extracts longer rationales than ISR, usually. However, it does not always mean that it carries supportive information for the outcome.}
    \label{fig: cont_582}
\end{figure*}

\begin{figure*}
    \begin{mdframed}[linewidth=.5pt, roundcorner=10pt, backgroundcolor=gray!1]
        \begin{subfigure}{1\textwidth}
            \small{
                ... on 23 july 1999 and 8 february 2000 . 7 . in the course of the proceedings the applicant discovered; in 1999 . the transfer of ownership had been formally registered on 3 march 1999 . on 24 february 2000 the applicant therefore; void . 8 . three hearings were adjourned because of the defendant [UNK] s absence . 9 . on 13 march 2001 the district court dismissed the applicant; flat . 10 . three hearings were again adjourned due to the defendant [UNK] s failure to appear . because of the defendant [UNK] s medical condition , a district court official heard the defendant at his home on 15 august 2002 . 11 . the judge dealing with the case was; ...
            }
            \caption{Concatenated rationales, which were extracted by MaRC}
            \label{subfig: cont_582_rat_1}
        \end{subfigure}
        \vspace{0.3cm}
        \begin{subfigure}{1\textwidth}
            \small{ 
                ... the case 4 . the applicant was born in 1979 and lives in kosice . a . proceedings concerning the purchase of a flat 5 . on 30 march 1999 the applicant filed a civil action with the kosice ii district court . she; appeal and , as a result , the kosice regional court discontinued the proceedings on 12 september 2003 . b . constitutional proceedings 15 . on 19 march 2003 the constitutional court concluded that the applicant [UNK] s right to a hearing; right to a hearing within a reasonable time .; ...
            }
            \caption{Concatenated rationales, which were extracted by ISR}
            \label{subfig: cont_341_rat_2}
        \end{subfigure}
        \vspace{0.3cm}
        \hfill
        \begin{subfigure}{1\textwidth}
            \small{
             ... On 30 March 1999 the applicant filed a civil action with the Košice II District Court. She claimed that the defendant should be ordered, in accordance with his earlier written undertaking, to sell a flat to her; In the course of the proceedings the applicant discovered that the defendant had transferred the ownership of the flat in issue to a different person in 1999. The transfer of ownership had been formally registered on 3 March 1999. On 24 February 2000 the applicant therefore requested that the new owner of the flat should join the proceedings as a defendant.; On 4 February 2003 the District Court delivered a judgment by which it dismissed the applicant’s action.; ...
            }            
        \caption{Concatenated rationales, which were extracted by expert A}
        \label{subfig: cont_582_rat_3}
        \end{subfigure}

    \end{mdframed}
    \caption{Example from concatenated rationales by MaRC (a), ISR (b) and expert A (c), from the case of \href{https://hudoc.echr.coe.int/eng?i=001-92582}{Buľková v. Slovakia, 2017}. Expert B assesses global support and sufficiency criteria using such documents.}
    \label{fig: cont_582_rat}
\end{figure*}

\section{LLM-as-a-Judge evaluations} \label{sec: app_prompts}
\subsection{Prompts for LLM-as-a-Judge setting}


To use LLMs as a tool for qualitative analysis, we adopt an expert-angle prompt design \cite{parfenova_text_2025}. Figures~\ref{fig: prompt_supp_single_shot} and~\ref{fig: prompt_suff_single_shot} show the single-shot LLM-as-a-Judge templates for the global rationale support and sufficiency criteria, respectively. We conducted several pilot experiments and refined these prompts iteratively.

We observed that asking only for a label and a confidence level does not adequately reveal the LLMs' reasoning for interpretability evaluation. Therefore, we also ask LLMs to provide explanations for both their answer and their stated confidence. In the few-shot setting, we include example cases (Figure \ref{fig: few_shot_examples}) such that the examples and the target case all involve alleged violations of the same article.

For these experiments we employ following LLMs:
\begin{itemize}
    \item \textbf{LLAMA}: LLAMA-3.1-70B-Instruct model by META can be accessed from the following link: \url{https://huggingface.co/meta-llama/Llama-3.1-70B-Instruct};
    \item \textbf{Mistral}: Mixtral-7$\times$8B-Instruct-v0.1 model by MistralAI can be accessed via following link: \url{https://huggingface.co/mistralai/Mixtral-8x7B-Instruct-v0.1};
    \item \textbf{Gemma}: Gemma-2-27b-it by Google, can be accessed from the following link: \url{https://huggingface.co/google/gemma-2-27b-it}.
\end{itemize}

In the single-shot setting we use the prompts from Figures~\ref{fig: prompt_supp_single_shot} and~\ref{fig: prompt_suff_single_shot}, directly. These prompt templates can be grouped into two parts: system content (RULES, CONFIDENCE RULES and OUTPUT SCHEMA) and user content (ARTICLE and RATIONALES). Because we use different models, the configuration varies due to the chat templates of these models. Llama support system and user roles in the single-shot setting,  whereas Mixtral and Gemma do not have a system role.  Therefore, we provide both groups of content in a single user message (see Figure \ref{fig: prompting_strategy}) for Mixtral and Gemma models.
\begin{figure*}[h!]
    \begin{mdframed}
        \small{

            \texttt{You are a strict legal evaluation assistant. Follow every rule exactly.}\\[6pt]
            
            \texttt{\textbf{RULES:}}\\
            \texttt{1) Use ONLY the provided ARTICLE and RATIONALE texts to reason internally.}\\
            \texttt{Do NOT use external legal knowledge or assumptions.}\\
            \texttt{2) DO NOT repeat, reprint, or serialize the RATIONALES text in your output.}\\
            \texttt{Your output MUST NOT contain the RATIONALES content.}\\
            \texttt{3) Decide whether the provided RATIONALES, taken together, SUPPORT or UNSUPPORT}\\
            \texttt{the claim that the ARTICLE applies to the CASE.}\\
            \texttt{SUPPORT = the rationales collectively establish a connection between}\\
            \texttt{the ARTICLE and the facts of the CASE.}\\
            \texttt{UNSUPPORT = the rationales fail to establish such a connection,}\\
            \texttt{or actively contradict it.}\\
            \texttt{4) Output ONLY one valid JSON object.}\\
            \texttt{No preamble, no trailing text, no markdown fences.}\\
            \texttt{5) Keep explanation to 1--3 sentences.}\\
            \texttt{Do NOT include full case text in explanations.}\\
            \texttt{Cite rationales by ID in square brackets, e.g. [R1] or [R1,R2].}\\[6pt]
            
            \texttt{\textbf{CONFIDENCE RULES:}}\\
            \texttt{- confidence reflects how certain you are about your own}\\
            \texttt{~~SUPPORT / UNSUPPORT decision --- nothing else.}\\
            \texttt{- It does NOT reflect how certain you are that a violation occurred in reality.}\\
            \texttt{- It does NOT reflect how strongly the Article applies to the case in general.}\\
            \texttt{- Ask yourself: "How certain am I that these rationales SUPPORT or UNSUPPORT}\\
            \texttt{~~the Article-Case connection?" That is your confidence.}\\[4pt]
            \texttt{- HIGH~~: You are certain about your decision. The rationales clearly and}\\
            \texttt{~~~~~~~~~~~unambiguously point to one conclusion. A different reader would}\\
            \texttt{~~~~~~~~~~~almost certainly reach the same conclusion.}\\
            \texttt{- MEDIUM: You are reasonably sure about your decision but can see how a}\\
            \texttt{~~~~~~~~~~~different reader might disagree. Some ambiguity exists in the rationales.}\\
            \texttt{- LOW~~~: You are uncertain about your decision.}\\
            }
            \texttt{\textbf{OUTPUT SCHEMA} (reproduce this structure exactly):}\\
            \texttt{\{}\\
            \texttt{~~"decision":~~~~"<SUPPORT | UNSUPPORT>",}\\
            \texttt{~~"confidence":~~"<LOW | MEDIUM | HIGH>",}\\
            \texttt{~~"explanation":~"<1--3 sentences citing [R\#] only>"}\\
            \texttt{\}}\\[6pt]
            
            \texttt{\textbf{ARTICLE:}}\\
            \texttt{\{article\_text\}}\\[4pt]
            \texttt{\textbf{RATIONALES:}}\\
            \texttt{\{rationales\_text\}}\\[4pt]
            \texttt{Now evaluate and output the JSON.}
    \end{mdframed}
    \caption{Prompt is designed for global rationale support criterion for the LLMs-as-a-Judge with a single-shot setting. Notice that Article X, can be either Article 6 or 8, including their bullet-points.}
    \label{fig: prompt_supp_single_shot}
\end{figure*}

\begin{figure*}[h!]
    \begin{mdframed}
        \small{
            \texttt{You are a strict legal evaluation assistant. Follow every rule exactly.}\\[6pt]
            
            \texttt{\textbf{RULES:}}\\
            \texttt{1) Use ONLY the provided ARTICLE and RATIONALE texts to reason internally.}\\
            \texttt{Do NOT use external legal knowledge or assumptions.}\\
            \texttt{2) DO NOT repeat, reprint, or serialize the RATIONALES text in your output.}\\
            \texttt{Your output MUST NOT contain the RATIONALES content.}\\
            \texttt{3) Decide whether the provided RATIONALES, taken together, are SUFFICIENT or INSUFFICIENT}\\
            \texttt{to deduce that a violation of the ARTICLE occurred in the CASE.}\\
            \texttt{SUFFICIENT = the rationales collectively provide enough evidence}\\
            \texttt{to conclude that a violation occurred.}\\
            \texttt{INSUFFICIENT = the rationales do not provide enough evidence}\\
            \texttt{to reach such a conclusion.}\\
            \texttt{4) Output ONLY one valid JSON object.}\\
            \texttt{No preamble, no trailing text, no markdown fences.}\\
            \texttt{5) Keep explanation to 1--3 sentences.}\\
            \texttt{Do NOT include full case text in explanations.}\\
            \texttt{Cite rationales by ID in square brackets, e.g. [R1] or [R1,R2].}\\[6pt]
            
            \texttt{\textbf{CONFIDENCE RULES:}}\\
            \texttt{- confidence reflects how certain you are about your own}\\
            \texttt{~~SUFFICIENT / INSUFFICIENT decision --- nothing else.}\\
            \texttt{- It does NOT reflect how certain you are that a violation occurred in reality.}\\
            \texttt{- It does NOT reflect how strongly the Article applies to the case in general.}\\
            \texttt{- Ask yourself: "How certain am I that these rationales are SUFFICIENT or INSUFFICIENT}\\
            \texttt{~~to establish the violation?" That is your confidence.}\\[4pt]
            \texttt{- HIGH~~: You are certain about your decision. The rationales clearly provide}\\
            \texttt{~~~~~~~~~~~enough --- or clearly do not provide enough --- evidence to deduce}\\
            \texttt{~~~~~~~~~~~the violation. A different reader would almost certainly reach the same conclusion.}\\
            \texttt{- MEDIUM: You are reasonably sure about your decision but can see how a}\\
            \texttt{~~~~~~~~~~~different reader might disagree. The evidence is present but incomplete}\\
            \texttt{~~~~~~~~~~~or partially ambiguous.}\\
            \texttt{- LOW~~~: You are uncertain about your decision. The rationales are too sparse,}\\
            \texttt{~~~~~~~~~~~fragmented, or contradictory to confidently assess whether they are}\\
            \texttt{~~~~~~~~~~~sufficient or not.}\\[6pt]
            
            \texttt{\textbf{OUTPUT SCHEMA} (reproduce this structure exactly):}\\
            \texttt{\{}\\
            \texttt{~~"decision":~~~~"<SUFFICIENT | INSUFFICIENT>",}\\
            \texttt{~~"confidence":~~"<LOW | MEDIUM | HIGH>",}\\
            \texttt{~~"explanation":~"<1--3 sentences citing [R\#] only>"}\\
            \texttt{\}}\\[6pt]
            
            \texttt{\textbf{ARTICLE:}}\\
            \texttt{\{article\_text\}}\\[4pt]
            \texttt{\textbf{RATIONALES:}}\\
            \texttt{\{rationales\_text\}}\\[4pt]
            \texttt{Now evaluate and output the JSON.}
        }        
    \end{mdframed}
    \caption{Prompt is designed for global rationale sufficiency criterion for the LLMs-as-a-Judge with a single-shot setting. Notice that Article X, can be either Article 6 or 8, including their bullet-points.}
    \label{fig: prompt_suff_single_shot}
\end{figure*}
For the few-shot setting, we use Expert B’s evaluations as examples. Following a leave-one-out strategy, we provide the LLM with Expert B’s answers for the other cases involving the same article, excluding the case currently being evaluated. Figure~\ref{fig: few_shot_examples} illustrates this few-shot configuration.
\begin{figure*}
    \begin{mdframed}
        \small{
            \texttt{\textbf{EXAMPLE 1}}\\
            
            \texttt{\textbf{ARTICLE:}}\\
            \texttt{\{article\_text\}}\\[4pt]
            \texttt{\textbf{RATIONALES:}}\\
            \texttt{\{rationales\_text\}}\\[4pt]

            \texttt{\textbf{EXPERT B:}}\\
            \texttt{\{}\\
            \texttt{~~"decision":~~~~"<SUFFICIENT | INSUFFICIENT>",}\\
            \texttt{~~"confidence":~~"<LOW | MEDIUM | HIGH>",}\\
            \texttt{~~"explanation":~"<1--3 sentences citing [R\#] only>"}\\
            \texttt{\}}

            $\vdots$

            $\vdots$

            \texttt{\textbf{EXAMPLE 4}}\\
            
            \texttt{\textbf{ARTICLE:}}\\
            \texttt{\{article\_text\}}\\[4pt]
            \texttt{\textbf{RATIONALES:}}\\
            \texttt{\{rationales\_text\}}\\[4pt]

            \texttt{\textbf{EXPERT B:}}\\
            \texttt{\{}\\
            \texttt{~~"decision":~~~~"<SUFFICIENT | INSUFFICIENT>",}\\
            \texttt{~~"confidence":~~"<LOW | MEDIUM | HIGH>",}\\
            \texttt{~~"explanation":~"<1--3 sentences citing [R\#] only>"}\\
            \texttt{\}}

        }        
    \end{mdframed}
    \caption{The template for the examples to be used in the few-shot setting. The "EXAMPLE \#" and "EXPERT B" phrases are not part of the prompt. They are used to improve the readability of the flow for the reader.}
    \label{fig: few_shot_examples}
\end{figure*}

Since every model supports an assistant role, we present the example cases as user messages and Expert B’s evaluations as assistant messages. As in the single-shot setting, Mixtral and Gemma do not expose a separate system role. For these models, we merge the system content with the first example message, followed by the remaining assistant–user dialogue (see Figure~\ref{fig: prompting_strategy}).
\begin{figure*}
    \centering
    \begin{mdframed}
        \begin{subfigure}{0.48\textwidth}
            \begin{tikzpicture}[every node/.style={yslant=0, xslant=0},font=\sffamily\scriptsize]
                \definecolor{spring}{RGB}{173, 198, 224}
                \definecolor{azure}{RGB}{198, 239, 206}
                \definecolor{fuschia}{RGB}{255, 200, 60}
                \definecolor{limey}{RGB}{235, 120, 120}

                \begin{scope}[local bounding box=leftside]
                    \node[draw=none, rounded corners=6pt, fill=spring,
                        minimum width=3.2cm, minimum height=1.2cm,
                        align=center] (topL) at (0,0) { \small{\textbf{System message:}}\\[1pt]\texttt{\small{RULES}}\\\texttt{\small{CONFIDENCE RULES}}\\\texttt{\small{OUTPUT SCHEMA}}};
                    \node[draw=none, rounded corners=6pt, fill=azure,
                        minimum width=3.2cm, minimum height=0.7cm,
                        below=0.1cm of topL, align=center] (bottomL)
                        {\textbf{\small{User message:}}\\[1pt]\texttt{\small{ARTICLE}}\\\texttt{\small{RATIONALES}}};
                    \node[draw=none, rounded corners=6pt, fill=fuschia,
                        minimum width=3.2cm, minimum height=0.7cm,
                        below=0.7cm of bottomL, align=center] (llama)
                        {\small{Llama}};
                \end{scope}

                \begin{scope}[local bounding box=rightside]
                    \node[draw=none, rounded corners=6pt, fill=azure,
                        minimum width=3.2cm, minimum height=1.9cm,
                        right=2.3cm of leftside.west,
                        align=center]  at (0,-0.35) (rightBox)
                        {\textbf{\small{User message:}}\\[1pt]\texttt{\small{RULES}}\\\texttt{\small{CONFIDENCE RULES}}\\\texttt{\small{OUTPUT SCHEMA}}\\[4pt]
                        \texttt{\small{ARTICLE}}\\\texttt{\small{RATIONALES}}};    
                    \node[draw=none, rounded corners=6pt, fill=limey,
                        minimum width=3.2cm, minimum height=0.7cm,
                        below=1.2cm of rightBox, align=center] (gemma)
                        {\small{Mixtral / Gemma}};
                    \end{scope}
                
                \draw[->, thick, black] ([yshift=0.0cm]bottomL.south) 
                  -- ([yshift=-0.7cm, xshift=0cm]bottomL.south) 
                  node[above, font=\scriptsize, midway] {};
                
                \draw[->, thick, black] ([yshift=0.0cm]rightBox.south) 
                  -- ([yshift=-1.2cm, xshift=0cm]rightBox.south) 
                  node[above, font=\scriptsize, midway] {};

            \end{tikzpicture}
            \caption{}
            \label{fig: prompting_strategy_a}
        \end{subfigure}
        \begin{subfigure}{0.4\textwidth}
            \begin{tikzpicture}[font=\sffamily\small, >=stealth, node distance=0.8cm]

                \definecolor{spring}{RGB}{173, 198, 224}
                \definecolor{azure}{RGB}{185, 165, 225}
                \definecolor{azure2}{RGB}{173, 198, 224}
                
                \definecolor{azure3}{RGB}{198, 239, 206}

                \definecolor{ochre}{RGB}{240, 145, 145}    
                \definecolor{dustyrose}{RGB}{140, 140, 140}

                \begin{scope}
                    \node[draw=none, rounded corners=6pt, fill=spring, minimum width=2.2cm, minimum height=0.6cm, align=center] (sysL) at (0,1.5) {\small{\texttt{System}}};
                    \node[draw=none, rounded corners=6pt, fill=azure3, minimum width=2.2cm, minimum height=0.6cm, align=center, below=0.2cm of sysL] (user1L) {\texttt{\small{Example 1 (User)}}};
                    \node[draw=none, rounded corners=6pt, fill=azure, minimum width=2.2cm, minimum height=0.6cm, align=center, below=0.2cm of user1L] (asst1L) {\texttt{\small{Expert B {\texttt{(Assistant)}}}}};
                    \node[draw=none, rounded corners=6pt, fill=azure3, minimum width=2.2cm, minimum height=0.6cm, align=center, below=0.2cm of asst1L] (user2L) {\texttt{\small{Example 2 (User)}}};
                    \node[draw=none, rounded corners=6pt, fill=azure, minimum width=2.2cm, minimum height=0.6cm, align=center, below=0.2cm of user2L] (asst2L) {\texttt{\small{Expert B (Assistant)}}};
                    \node[draw=none, rounded corners=6pt, fill=azure3, minimum width=2.2cm, minimum height=0.6cm, align=center, below=0.2cm of asst2L] (user3L) {\texttt{\small{User}}};
                \vdots
                \end{scope}

                \draw[thick, ->] (sysL.south) -- (user1L.north);
                \draw[thick, ->] (user1L.south) -- (asst1L.north);
                \draw[thick, ->] (asst1L.south) -- (user2L.north);
                \draw[thick, ->] (user2L.south) -- (asst2L.north);
                \draw[thick, ->] (asst2L.south) -- (user3L.north);
                
                \begin{scope}
                \node[draw=none, rounded corners=6pt, fill=dustyrose, minimum width=3.2cm, minimum height=1.2cm, align=center, right=3cm of sysL, anchor=north] (user1R) {
                \colorbox{spring}{\makebox[2.4cm][c]{\texttt{\small{System}}}}\\[5pt]
                
                \colorbox{azure3}{\makebox[2.4cm][c]{\textcolor{black}{\texttt{\small{Example 1 (User)}}}}}\\
                };
                \node[draw=none, rounded corners=6pt, fill=azure, minimum width=2.2cm, minimum height=0.6cm, align=center, below=0.2cm of user1R] (asst1R) {\texttt{\small{Expert B (Assistant)}}};
                \node[draw=none, rounded corners=6pt, fill=azure3, minimum width=2.2cm, minimum height=0.6cm, align=center, below=0.2cm of asst1R] (asst2R) {\texttt{\small{Expert B (Assistant)}}};
                \node[draw=none, rounded corners=6pt, fill=azure, minimum width=2.2cm, minimum height=0.6cm, align=center, below=0.2cm of asst2R] (user2R) {\texttt{\small{Expert B (Assistant)}}};                    
                \end{scope}

                \draw[thick, ->] (user1R.south) -- (asst1R.north);
                \draw[thick, ->] (asst1R.south) -- (asst2R.north);
                \draw[thick, ->] (asst2R.south) -- (user2R.north);
                
            \end{tikzpicture}
            \caption{}
            \label{fig: prompting_strategy_b}
        \end{subfigure}

    \end{mdframed}
    \caption{Prompt templates for the single- (a) and few-shot (b) settings. In the few-shot setting we use 4 cases as examples, unlike as given in this Figure (b). }
    \label{fig: prompting_strategy}
\end{figure*}

We also experimented with SaulLM-54B-Instruct \cite{colombo2024saullm}, given its legal pre-training. However, the model does not provide a dedicated chat template, so we could not use it reliably in our few-shot setting. We manually inserted Mixtral’s chat template, since SaulLM belongs to the same family, but this did not resolve the issue. In addition, we observed that SaulLM occasionally appeared to leak pre-training data into its answers.

\subsection{LLM-as-a-Judge results}

In Figures \ref{fig: support_malayevy} and \ref{fig: sufficiency_malayevy} we show explanations of expert B and LLMs for their evaluation results of rationales extracted from the case of \href{https://hudoc.echr.coe.int/eng?i=001-175482}{Malayevy v. Russia} by ISR, which are depicted in Figure \ref{fig: rat_snippet}, respectively. In answers we follow the format of [Answer]: [Explanation]. 

\begin{figure*}
    \begin{mdframed}[linewidth=.5pt, roundcorner=10pt, backgroundcolor=gray!1]
        \small{
        \begin{tabular}{lm{2.5cm} m{10.5cm}}
            \textbf{Expert} &\textbf{Expert B} & \vspace{0.1cm}[UNSUPPORT]: Random words, rationales do not make any sense at all. \vspace{0.1cm}\\
            \midrule
            \multirow{3}{*}{\textbf{Llama}} &\textbf{Single-shot} & \vspace{0.1cm}[UNSUPPORT]:  The rationales provided do not establish a clear connection between the article and the case, as they appear to be fragmented and lack a coherent narrative [r1-r25]. \\
            \cmidrule{2-3}
            &\textbf{Few-shot, informal} & \vspace{0.1cm} [UNSUPPORT]: The rationales do not provide any information about the violation of the right to respect for private and family life \\
            \cmidrule{2-3}
            
            & \textbf{Few-shot, formal} & \vspace{0.1cm} [UNSUPPORT]: The rationales do not provide sufficient information to establish a connection to the article, and appear to be primarily focused on property transactions and disputes. \\
            \midrule
            \multirow{3}{*}{\textbf{Mixtral}} &\textbf{Single-shot} & \vspace{0.1cm} [UNSUPPORT]: The rationales do not provide sufficient information about any potential interference by a public authority with the right to respect for private and family life, home, or correspondence, nor is there any mention of such an authority's actions being in accordance with the law and necessary in a democratic society [r1-r25].\\
            \cmidrule{2-3}
            
            &\textbf{Few-shot, informal} & \vspace{0.1cm}[UNSUPPORT]: The rationales do not provide enough information to make a decision. they are not clear and do not contain any information about the violation of the article 8\\
            \cmidrule{2-3}
            
            &\textbf{Few-shot, formal} & \vspace{0.1cm} [UNSUPPORT]: The rationales do not provide sufficient information to establish a connection to the article. they are largely incoherent and do not indicate a possible violation.\\
            \midrule
            \multirow{3}{*}{\textbf{Gemma}} & \textbf{Single-shot} & \vspace{0.1cm}[UNSUPPORT]: The rationales describe a property dispute and eviction proceedings, not interference with private or family life as defined in article 8 [r1, r2, r22].\\
            \cmidrule{2-3}
            
            &\textbf{Few-shot, formal}& \vspace{0.1cm}[UNSUPPORT]: The rationales are largely incoherent and do not provide sufficient information to establish a connection to the article.\\
            \cmidrule{2-3}            
            &\textbf{Few-shot, informal}&\vspace{0.1cm}[UNSUPPORT]: The rationales are very fragmented and do not provide a clear picture of the situation. it is impossible to determine if there was a violation of article 8 based on these.\\
        \end{tabular}
        }
    \end{mdframed}
    \caption{Comparative analysis of Expert B’s explanations and LLMs’ explanations (in single- and few-shot settings) for the global rationale support criterion in the case \href{https://hudoc.echr.coe.int/eng?i=001-175482}{Malayevy v. Russia}, based on rationales extracted by ISR. We observe that all LLMs are aligned with Expert B on their decisions about the rationale snippets in this document (see Figure~\ref{fig: rat_snippet} for the rationales).}
    \label{fig: support_malayevy}
\end{figure*}

\begin{figure*}
    \begin{mdframed}[linewidth=.5pt, roundcorner=10pt, backgroundcolor=gray!1]
        \small{
        \begin{tabular}{lm{2.5cm} m{10.5cm}}
            \textbf{Expert} &\textbf{Expert B} & \vspace{0.1cm}[INSUFFICIENT]: Not sufficient at all. just random words. nothing can be extracted from these and i would need "normal" rationales to decide anything \vspace{0.1cm}\\
            \midrule
            \multirow{3}{*}{\textbf{Llama}} &\textbf{Single-shot} & \vspace{0.1cm}[INSUFFICIENT]: The rationales provided are fragmented and lack clear connections to the article, making it difficult to determine if a violation occurred [r1-r25]. \\
            \cmidrule{2-3}
            &\textbf{Few-shot, informal} & \vspace{0.1cm} [INSUFFICIENT]: The rationales are very short and lack context, making it impossible to understand the case and determine if there was a violation of article 8. more information is needed to make a decision \\
            \cmidrule{2-3}
            
            & \textbf{Few-shot, formal} & \vspace{0.1cm} [INSUFFICIENT]: The rationales are largely incoherent and lack a clear narrative, making it impossible to establish the basis for a violation.\\
            \midrule
            \multirow{3}{*}{\textbf{Mixtral}} &\textbf{Single-shot} & \vspace{0.1cm}[INSUFFICIENT]: The rationales do not provide enough information about the nature of the transactions, agreements, or disputes between the parties to determine if there was a violation of the right to respect for private and family life [r1-r10, r13-r18, r20, r22-r25].\\
            \cmidrule{2-3}
            
            &\textbf{Few-shot, informal} & \vspace{0.1cm}[INSUFFICIENT]:  The rationales are not sufficient at all. they are just a bunch of random facts and there is no clear story or any kind of explanation of what happened. would need a lot more context and information to be able to decide on the violation\\
            \cmidrule{2-3}
            
            &\textbf{Few-shot, formal} & \vspace{0.1cm} [INSUFFICIENT]: The rationales are incomplete and lack coherence, making it impossible to establish the basis for a violation.\\
            \midrule
            \multirow{3}{*}{\textbf{Gemma}} & \textbf{Single-shot} & \vspace{0.1cm}[INSUFFICIENT]: The rationales describe a property dispute and legal proceedings but do not clearly establish interference with private or family life by a public authority [r1, r2, r22].\\
            \cmidrule{2-3}
            
            &\textbf{Few-shot, formal}& \vspace{0.1cm}[INSUFFICIENT]: The rationales are fragmented and lack context. the information provided is insufficient to establish the basis for a violation of article 8. \\
            \cmidrule{2-3}            
            &\textbf{Few-shot, informal}&\vspace{0.1cm}[INSUFFICIENT]: The rationales are very fragmented and lack context. it's impossible to understand the situation or the alleged violation of article 8 based on these snippets\\
        \end{tabular}
        }
    \end{mdframed}
    \caption{Comparative analysis of Expert B’s explanations and LLMs’ explanations (in single- and few-shot settings) for the global rationale sufficiency criterion in the case \href{https://hudoc.echr.coe.int/eng?i=001-175482}{Malayevy v. Russia}, based on rationales extracted by ISR. We observe that all LLMs are aligned with Expert B on their decisions about the rationale snippets in this document (see Figure~\ref{fig: rat_snippet} for the rationales).}
    \label{fig: sufficiency_malayevy}
\end{figure*}

\begin{table*}[]
\setlength{\tabcolsep}{3pt}
\resizebox{\textwidth}{!}{%
\begin{tabular}{l| rrr | rrr}
\toprule
Pair & \multicolumn{3}{c|}{Sufficiency} & \multicolumn{3}{c}{Support} \\
\cmidrule(lr){2-4}\cmidrule(lr){5-7}
     & Single-shot & Few-shot & Few-shot (formal) & Single-shot & Few-shot & Few-shot (formal)\\
\midrule
Gemma-Llama
  & $0.27$ $[-0.04, 0.57]$ & $0.4$ $[0.18, 0.67]$ & $0.59$ $[0.30, 0.86]$
  & $\mathbf{0.19}$ $[-0.18, 0.53]$ & $\mathbf{0.52}$ $[0.09, 0.84]$ & $\mathbf{0.17}$ $[-0.13, 0.62]$\\
Mixtral-Gemma
  & $0.19$ $[-0.17, 0.56]$ & $\mathbf{0.52}$ $[0.15, 0.83]$ & $\mathbf{0.87}$ $[0.67, 1.00]$
  & $0.10$ $[0.00, 0.29]$ & $0.08$ $[0.00, 0.21]$ & $0.14$ $[0.03, 0.31]$\\
Mixtral-Llama
  & $\mathbf{0.27}$ $[0.07, 0.53]$ & $0.4$ $[0.17, 0.67]$ & $0.48$ $[0.23, 0.74]$
  & $0.10$ $[0.00, 0.29]$ & $0.04$ $[0.00, 0.12]$ & $0.04$ $[0.00, 0.14]$\\
\bottomrule
\end{tabular}
}
\caption{Inter-LLM Cohen's $\kappa$ (95\% CI): Pairwise agreement for Sufficiency (left) vs. Support (right)}
\label{tab: llm-llm}
\end{table*}

\end{document}